\documentclass{article}
\usepackage{fullpage}
\usepackage{natbib}
\usepackage[utf8]{inputenc}  
\usepackage[T1]{fontenc}    
\usepackage{hyperref}    
\usepackage{url}           
\usepackage{booktabs}      
\usepackage{amsfonts}       
\usepackage{nicefrac}       
\usepackage{microtype}      
\usepackage{xcolor}       
\usepackage{graphicx}
\usepackage{amsmath}
\usepackage{algorithm}
\usepackage{algorithmicx}
\usepackage{algpseudocode}
\usepackage{subfigure}
\usepackage{amssymb}
\DeclareMathOperator*{\argmax}{arg\,max}
\DeclareMathOperator*{\argmin}{arg\,min}

\title{Generative modeling of density regression through tree flows}

\author{Zhuoqun Wang\\
Department of Statistical Science, Duke University\\
\texttt{zw122@duke.edu}\\\\
 Naoki Awaya \\
School of Political Science and Economics,  Waseda University\\
\texttt{nawaya@waseda.jp}\\\\
Li Ma\\
Department of Statistical Science, Duke University\\
\texttt{li.ma@duke.edu}
}

\begin{document}

\maketitle

\begin{abstract}
  
A common objective in the analysis of tabular data is estimating the conditional distribution (in contrast to only producing predictions) of a set of ``outcome'' variables given a set of ``covariates'', which is sometimes referred to as the ``density regression'' problem. Beyond estimation on the conditional distribution, the generative ability of drawing synthetic samples from the learned conditional distribution is also desired as it further widens the range of applications. We propose a flow-based generative model tailored for the density regression task on tabular data. Our flow  applies a sequence of tree-based piecewise-linear transforms on initial uniform noise to eventually generate samples from complex conditional densities of (univariate or multivariate) outcomes given the covariates and allows efficient analytical evaluation of the fitted conditional density on any point in the sample space. We introduce a training algorithm for fitting the tree-based transforms using a divide-and-conquer strategy that transforms maximum likelihood training of the tree-flow into training a collection of binary classifiers---one at each tree split---under cross-entropy loss. We assess the performance of our method under out-of-sample likelihood evaluation and compare it with a variety of state-of-the-art conditional density learners on a range of simulated and real benchmark tabular datasets. Our method consistently achieves comparable or superior performance at a fraction of the training and sampling budget. Finally, we demonstrate the utility of our method's generative ability through an application to generating synthetic longitudinal microbiome compositional data based on training our flow on a publicly available microbiome study. 
\end{abstract}

\section{Introduction}

Many data analytical tasks involving tabular data require learning the conditional distribution of a (possibly multivariate) outcome $y$ given a set of contextual variables (or covariates) $x$, and  generating new observations of $y$ conditional on the value of $x$. Given the effectiveness of tree-ensemble based approaches for characterizing tabular data in both supervised learning \citep{grinsztajn2022treebased} and generative modeling of joint (i.e., unconditional) multivariate distributions \citep{iouyue2018ddd, awaya2023}, we aim to introduce an efficient approach to approximate conditional densities of tabular data using ensembles of tree-based transforms. Specifically, we introduce a tree-based normalizing flow capable of (1) outputting the fitted density $p(y|x)$ for any given pair of value $(x,y)$; (2) efficiently generating $y$ given $x$ from the estimated distribution; and (3) being trained efficiently based on maximum likelihood using tree-fitting algorithms that requires a computational budget linear in the sample size.

Key to efficient training of our tree flow is a new single-tree learning algorithm for approximating conditional densities, which is employed repeatedly to find the sequence of tree-based transforms whose composition constitutes the flow. The single-tree learning algorithm transforms the unsupervised problem of learning a conditional density into a collection of supervised problems involving binary classification, one at each tree split, and accomplishes maximum likelihood fitting on the tree-based transform through minimizing the cross-entropy loss on the corresponding binary classification tasks. The framework allows any binary classifier, and in particular non-tree based classifiers to be incorporated, thereby complementing the effectiveness of the tree-based transforms in approximating the conditional density. 
For illustration, we assess the performance of the resulting tree-flow using logistic regression and multi-layper perceptrons (MLPs) as the binary classifier at the tree splits.

The tree-based transforms are all piecewise linear mappings with closed form expressions derived from the trained binary classifiers, and so are their inverses. The Jacobian of the piecewise linear transforms is piecewise constant and corresponds exactly to the fitted conditional density at each iteration of the tree fitting to the current ``observations'' and therefore are available immediately as an output of the tree fitting algorithms during training. The sampling stage of the algorithm involves simply applying a sequence of piecewise linear transforms, which are the inverses of the transforms learned during training, to uniform noise which can be carried out in linear time.

In summary, we propose a flow-based generative model for conditional density $p(y|x)$ that utilizes tree-based transforms along with covariate-dependent probability splits to approximate the conditional density, while offering exact density evaluation and efficient training and sampling. 
Some unique features of our method are
\begin{itemize}

    \item \textbf{Combining the strength of trees and non-tree based approximations.} Our approach exploits the effectiveness of tree-based transforms in approximating multivariate distribution on tabular data, along with the additional flexibility of non-tree based approximators such as logistic regression and neural network (NN)-based binary classifiers to approximate the smooth varying density over covariate values. We empirically show that the tree and non-tree hybrid approach can achieve superior performance on conditional density estimation tasks involving tabular data over other state-of-the-art methods based only on NNs.  

    \item \textbf{Efficient training and sampling}. We employ a divide-an-conquer strategy by converting the unsupervised density learning problem to a collection of  binary classification problems defined on the tree splits, and introduce a tree-fitting algorithm with $O(ndq)$ time complexity for a training set of $n$ observations with $d$ outcome variables and $q$ conditioning covariates. 
    Sampling from the trained flow involves applying a sequence of piecewise linear transforms to uniform noise, which can be completed efficiently at complexity $O(q)$ for drawing each sample, independent of $d$.
\end{itemize}

Because our approach falls into the general class of normalizing flows (NFs), it also inherits the general desirable properties of NFs, including

\begin{itemize}

\item \textbf{Exact conditional density evaluation}. Our method allows evaluating the fitted conditional density at any point in the sample space. In particular, the time complexity of evaluating the conditional density of one sample with our method is $O(q)$, independent of the number of outcome variables. 

\end{itemize}

We carry out extensive numerical experiments to assess the performance of our method in density estimation and compare it to a range of state-of-the-art competitors on both simulated and real benchmark datasets. We consider an application to a longitudinal microbiome compositional dataset in which we generate synthetic microbiome compositions given time as a covariate. The results showcase the effectiveness of our method in capturing complex multivariate conditional densities.

\section{A conditional flow with tree-based transforms}
\label{sec:method}

\subsection{A tree ensemble-based approximation to conditional densities}
The problem of conditional density estimation is to find a close approximation $f_x: \mathcal{Y}\rightarrow R$ to an unknown conditional distribution $p(\cdot|x)$ given a training set of $n$ observations $\{(x_i,y_i)\}_{i=1}^n$, where $y_i|x_i\sim p(y_i|x_i)$ independently given the covariate $x_i$. We build the conditional distributions of a d-dimensional vector ${y}$ as normalizing flows. That is, given a set of covariate values $ x$, the vector $y$ can be obtained by applying a sequence of $x$-dependent invertible and differentiable transformations on a random variable $u$ uniformly distributed over the $d$-dimensional unit cube, $(0,1]^d$.

We train the normalizing flow to a target conditional distribution by maximum likelihood, i.e., minimizing the forward KL divergence. 
\cite{iouyue2018ddd} introduced a flow incorporating a class of piecewise linear transforms defined on binary partition trees. This class of tree-based flows was later more formally studied and shown to be analogous to ensemble tree approximators such as tree boosting by \cite{awaya2023}. The tree-based piecewise linear transforms generalize the notion of the cumulative distribution function (CDF) for a one-dimensional distribution to multivariate sample spaces based on a reordering of the sample space based on dyadic tree partition on the sample space. Accordingly, \cite{awaya2023} referred to this class of transforms as ``tree-CDF'' transforms.
In this work we continue to choose the tree-CDF transform as the basis for the flow model due to its computational advantage and expressive power established under the unconditional scenario \citep{awaya2023} as well as the evidence for the effectiveness of tree-ensemble based approximations to tabular data \citep{grinsztajn2022treebased}. Our first task is to generalize the tree-CDF to covariate-dependent tree-CDFs.

Without loss of generality, throughout this paper we assume the outcome observations are defined on the $d$-dimensional unit cube, that is, $y_i = (y_{i1},\cdots, y_{id}) \in (0,1]^d$, and we place no assumptions on the covariate space or distribution since the covariates $x_i$ will always be treated as given. 
Consider nested axis-aligned dyadic partitions on $(0,1]^d$ represented by a full dyadic tree $T$ with internal nodes $I(T)$ and leaf nodes $L(T)$. Each node of the tree represents a rectangular region resulted from the partitions. The root node is $(0,1]^d$, and each internal node is split into two children. 
Each finite dyadic tree $T$ gives rise to a piecewise constant conditional density given some covariate value $x$:
$g_x(y) = \sum_{A\in L(T)} c_{x,A}\,\mathbf{1}(y\in A)$, where $\mathbf{1}(\cdot)$ is the indicator function. The conditional density $g_x$ uniquely defines a conditional distribution for $y$ given $x$, denoted by $G_{x}$, and $g_x={\rm d}G_x/{\rm d}\mu$, where $\mu$ represents the Lebesgue measure.
Moreover, there exists a piecewise linear transform (which is invertible with analytic Jacobian) corresponding to the tree $T$ and the probabilities $c_{x,A}$, called the (covariate-dependent) ``tree-CDF'' and denoted by ${\bf G}_x:(0,1]^d\rightarrow (0,1]^d$ 
that generalizes the notion of univariate CDFs in the following sense:
$${\bf G}_x(y) \sim \text{Uniform}((0,1]^d) \text{ if } y\sim {g_x} \qquad \text{and} \qquad {\bf G}_x^{-1}(u) \sim g_x \text{ if } u\sim \text{Uniform}((0,1]^d).$$ 
Moreover, $|{\rm det} (J_{{\bf G}_x}(y))| = g_x(y)$. 
Further mathematical details of the tree-CDF are provided in Appendix~\ref{a:treecdf}. 
\cite{awaya2023} shows that compositions of tree-CDFs generalizes the notion of additive tree ensembles such as the one used in tree boosting for supervised learning to the unsupervised generative modeling context. 

Next we use tree-CDFs to construct our covariate-dependent normalizing flows. 
Specifically, to generate a sample $y$ from an arbitrary conditional probability distribution given some covariate $x$, we sample $u\sim {\rm Uniform}((0,1]^d)$, and then apply a sequence of transforms
\[
y = {\bf G}^{-1}_{1,x}\circ {\bf G}^{-1}_{2,x}\circ \cdots \circ {\bf G}^{-1}_{K,x}(u)
\]
where ${\bf G}^{-1}_{k,x}$ is the corresponding inverse of a tree-CDF which is also a piecewise linear mapping. Each tree-CDF in the sequence is associated with a distinct partition tree. The conditional distribution for $y$ is thus approximated by the additive ensemble of  single-tree conditional probability measures represented in the {\em group} formed by the tree-CDFs ${\bf G}_{k,x}$. (\cite{awaya2023} proves that the tree-CDFs and their inverses form a group in which the composition is the addition.)

By the chain rule, the log conditional density of $y$ is given by
\begin{align} \log f_x(y) = \sum_{k=1}^{K} \log g_{k,x}(y^{(k-1)})
\label{eq:logden}
\end{align}
where $g_{k,x}={\rm d}G_{k,x}/{\rm d}\mu$ is the corresponding piecewise constant density for $G_{k,x}$ with respect to the Lebesgue measure $\mu$, and $y^{(k-1)}={\bf G}_{k-1,x}\circ \cdots \circ {\bf G}_{1,x}(y)$ is called the ``{\em residual}'' at step $k$. (The notion of residuals is analogous to that in the supervised tree boosting, which serves in each step as the ``data'' for training the $k$th base learner, here $g_{k,x}$.) 

Eq.~\eqref{eq:logden} also implies that maximizing the likelihood, that is, finding the collection of densities $\{g_{k,x}:k=1,2,\ldots,K\}$ that maximizes $\sum_{i}\log f_{x_i}(y_i)$ based on training data $\{(x_i,y_i):i=1,2,\ldots,n\}$ can be achieved by iteratively maximizing the residual likelihood $\sum_{i} \log g_{k,x_i}(y_i^{(k-1)})$ over $g_{k,x}$ for $k=1,\dots,K$. See Algorithm~\ref{alg:ensemble} in Appendix~\ref{a:algo} for details. Next we address how to learn each $g_{k,x}$, or equivalently $G_{k,x}$ and ${\bf G}_{k,x}$ in detail.

\subsection{Fitting a single covariate-dependent tree-CDF through binary classification} 
\label{sec:single_tree}

The key to training the flow using Algorithm~\ref{alg:ensemble} (in Appendix~\ref{a:algo}) is the fitting of the individual (covariate-dependent) tree-CDF transform ${\bf G}_{k,x}$ based on the residuals $\{y_i^{(k-1)}:i=1,2,\ldots,n\}$. This involves learning the corresponding partition tree $T_k$ as well as the piecewise constant density $g_{k,x}$ defined on $T_k$. To this end, we introduce a divide-and-conquer strategy that efficiently accomplishes this task through transforming it into training a collection of binary classifications along the tree splitting decisions. This strategy also allows us to approximate the dependency of the outcome distribution on the covariates---which trees do not effectively approximate---through flexible approximators such as neural networks. 

First, we note that the conditional probability distribution $G_{k,x}$ can be expressed equivalently in terms of the probability it allocates at each tree split along the corresponding dyadic partition tree $T_k$ on the space of $y$. Specifically, 
for each internal node $A\in I(T_k)$ 
we let $A_l$ and $A_r$ be the two children nodes of $A$. Then $G_{k,x}(A_l|A)=G_{k,x}(A_l)/G_{k,x}(A)$ is the relative probability $G_{k,x}$ assigns to $A_l$ and similarly $G_{k,x}(A_r|A)=G_{k,x}(A_r)/G_{k,x}(A)$ that of $A_r$.
$G_{k,x}$ and thus $g_{k,x}$ is fully specified by these splitting probabilities $G_{k,x}(A_l|A)$ and $G_{k,x}(A_r|A)$ at all of the internal nodes of $T_k$.

Then we note that learning $G_{k,x}(A_l|A)$ can be viewed as a binary classification task on predicting whether an outcome $y$ in $A$ falls in $A_l$ or in $A_r$ given the corresponding covariate value $x$. As such, we can model $G_{k,x}(A_l|A)$ using any binary classifier
$${G}_{k,x}(A_l|A) =  p_{\theta_{k,A}}(x),$$
where the classification probability $p_{\theta_{k,A}}(x)$ is parametrized by $\theta_{k,A}$. In our later numerical experiments, we consider the logistic regression and the multi-layper perceptrons (MLPs) as well as a combination of the two as the binary classifier, though the choice of the binary classifier can really be up to the practitioner and different classifiers can be adopted on different nodes $A$.

Next we describe how to train both the tree $T_k$ and the classifiers at the internal nodes of $T_k$. 
We eliminate $k$ in all subscripts in the following to avoid overly cumbersome notation. Let $\theta$ denote the collection of all binary classifiers on the internal nodes of the tree $T$. That is, $\theta = \{\theta_A:A\in I(T)\}$, where $\theta_A$ is the binary classifier associated with an individual internal node $A$. 

As we show in in Appendix~\ref{a:single_tree}, the log-likelihood can be decomposed along the tree splits as follows
\begin{equation}    
l(T,\theta) := \sum_{i=1}^n \log g_{x_i}(y_i)
=\sum_{A\in I(T)}\Bigl(l_{A,{\rm bin}}(T,\theta_A) + C_A(T)\Bigr)
\label{eq:ll}
\end{equation}
where
$l_{A,{\rm bin}}(T,\theta_A)=\sum_{y_i\in A}\left(\mathbf{1}(y_i\in A_l)\,\log p_{\theta_{A}}(x_i) + \mathbf{1}(y_i\in A_r)\,\log(1- p_{\theta_{A}}(x_i))\right)$ is the cross-entropy loss of the binary classifier,
and $C_A(T)=-n(A_l)\log\frac{\mu(A_l)}{\mu(A)} - n(A_r)\log\frac{\mu(A_r)}{\mu(A)}$ with $\mu$ being the Lebesgue measure 
and $n(A_l)$ and $n(A_r)$ the number of observations $y_i$ in $A_l$ and $A_r$ respectively. (One can also incorporate a penalty on the complexity of the tree $T$ into $C_A(T)$ 
for further regularization, which we discuss in Appendix~\ref{a:single_tree}.)

It is most important to note that the term $C_A(T)$ does not depend on the binary classifier $\theta$ or $x$. 
This means that for each $A$ maximizing $l_{A,{\rm bin}}(T,\theta_A)+C_A(T)$ over $(T,\theta_A)$ can proceed in two steps: first maximizing over $\theta_{A}$ by training a binary classifier based on the cross-entropy loss under each candidate way of splitting $A$, and then, maximizing over the ways to split $A$ based on the minimum loss from the trained binary classifier. 

Specifically, we describe this two-step training inductively. Suppose the current tree and corresponding node-level parameters are $(T_{j-1}^*,\theta_{j-1}^*)$. (At initiation, $T_{0}^*$ has only the root node, and $\theta_{0}^*$ is an empty set. Suppose there are $M$ possible ways to divide a node $A$ of $T_{j-1}^*$, yielding $M$ candidates for the tree structure, $\{T_{j,1}, \cdots, T_{j,M}\}$. Then
\begin{itemize}
    \item[Step 1.] Given $T\in\{T_{j,1},\cdots, T_{j,M}\}$, train the optimal binary classifier, $\theta^*_A(T)$, which minimizes the cross-entropy loss $l_{A,{\rm bin}}(T,\theta_A)$: 
      $\quad \theta^*_A(T) = \argmin_\theta l_{A,{\rm bin}}(T,\theta_A).$
    \item[Step 2.] Choose $T_j^* = \argmax_{T\in\{T_{j,1},\cdots, T_{j,M}\}} \left(l_{A,{\rm bin}}(T,\theta^*_A(T))+C_A(T)\right)$ and set $\theta_j^* = \theta_{j-1}^*\bigcup \{\theta^*_A(T_j^*)\}$. 
    
\end{itemize}

In Algorithm~\ref{algo:single_tree} in Appendix~\ref{a:algo}, we summarize the full root-to-leaf tree learning algorithm that starts off with the root (whole sample space of $y$), and expand one split at a time using the above two-step updating. A node is no longer split when it either reaches a predefined maximum depth or the number of samples in the node falls below a specified threshold. See Algorithm~\ref{algo:single_tree} in Appendix~\ref{a:algo} for details.

 Suppose the training set consists of \( n \) samples, each with $d$ outcomes and $q$ covariates. Our algorithm for training the tree flow exhibits linear time complexity \( O(ndq) \). Both the density evaluation of a test sample and generating a new sample operate at a complexity of \( O(q) \), independent of $d$. Detailed analysis of the time complexity is in Appendix~\ref{a:time_complexity}. This will also be confirmed empirically in our numerical experiments.

\subsection{Additional technical improvements}
\label{sec:additional}
We incorporate two additional technical strategies in training the tree flow that can lead to substantial improvement in applications. The first strategy involves regularization through shrinkage, which ensures that each tree transform in the flow modifies the residual distribution only slightly to avoid overfitting, and adopts different rates of transform at different spatial scales. The second strategy addresses the limitation of axis-aligned partition in the tree fitting by ensembling over multiple (covariate-dependent) rotations of the training data. 

{\it 1. Regularization through scale-dependent learning rate and early stopping.}
To avoid overfitting and smooth the resulting probability measure, a small learning rate can be applied to each tree-CDF to shrink its corresponding probability measure towards the uniform distribution, thereby achieving regularization. Specifically, in our implementation, the regularization can be applied in a scale-specific fashion by specifying a learning rate for each tree node $A$ according to the size of the set $A$, as measured by the Lebesgue measure $\mu(A)$. Specifically, we incorporate the scale-dependent learning rate $\{c(A)\}_{A\in I(T_k)}$ by setting
    $G_{k,x}(A_l|A) = c(A)p^\theta_k(x)+ (1-c(A))\frac{\mu(A_l)}{\mu(A)}$
where $c(A)$ is defined as $c(A) = c_0(1+\log_2\mu(A))^{-\gamma}$. The constant
$c_0\in(0,1)$ controls the global shrinkage level, and $\gamma\geq 0$ controls the rate at which shrinkage intensifies as the node volume decreases. Specifically, a $\gamma$ of 0 applies uniform shrinkage across all nodes regardless of their volume, whereas a positive $\gamma$ results in increased shrinkage at smaller nodes, serving as a form of ``soft pruning".

The total number \( K \) of tree-CDFs is determined using early stopping, which halts the algorithm when the log-likelihood on a separate validation set does not increase for \( w \) consecutive iterations, where \( w \) is a predefined window size. Additionally, since tree-CDFs may utilize various types of binary classifiers, this early stopping criterion can be independently applied to each classifier type. For example, the algorithm might initially use the logistic regression for node-level probability assignments. If there is no improvements in log-likelihood, this indicates that the logistic regression may no longer be capturing additional distributional structure from the training data. At this point, the algorithm could switch to a more complex classifier, such as MLPs, to attempt to extract more refined distributional structures, using potentially fewer tree-CDFs but with more complicated node-level models. As we shall see later in the experiments, such combination of classifiers can improve the performance over a single classifier.

The full algorithm for training the conditional tree flow that includes scale-specific shrinkage and early stopping is summarized in Algorithm~\ref{alg:ensemble}. 

{\it 2. Rotation ensemble of tree flows}.
To alleviate the restrictions associated with axis-aligned partitions and enhance the approximative ability of our tree flow, we propose using an ensemble of conditional tree flows trained on multiple rotated versions of the training data. 
To this end, we rotate the $y_i$'s in the original training data to generate \(J\) distinct data sets, denoted as \(D_1, \cdots, D_J\), where 
\(D_j = \{(x_i, y_iR_j)\}\), \(R_j\) is a rotation matrix applied to each data set. The ensemble model is built by training individual conditional tree flows on rotated datasets and taking a weighted average of their conditional densities with covariate-dependent weights. Specifically, to maintain computational efficiency we adopt an adaptive binning strategy in constructing the weights. We partition the covariate space $\mathcal{X}$ into disjoint regions, where weights are constant in each region but can vary across regions. Within each region, the weights are determined by the likelihood of the rotated training samples—rotations that yield a higher likelihood are up-weighted. (The details of the weights are provided in {Appendix~\ref{a:rotation}}.)
In our experiments, equally spaced 2D rotations are used, and $\mathcal{X}$ is partitioned using k-means clustering, with further details provided in Section~\ref{sec:exp}.

\section{Experiments}
\label{sec:exp}

\subsection{Real-world tasks with univariate outcomes}
\label{sec:1dexp}
We first assess the performance of the proposed method in estimating the conditional density for univariate outcomes. This experiment involves nine benchmark datasets recorded in the University of California, Irvine (UCI) machine learning repository\citep{uci}, whose characteristics are summarized in Table~\ref{uci1} in Appendix~\ref{a: additional experimental results}. For each trial, the data are randomly split into a training set and a test set with a ratio of 9:1. We use the same preprocessed data as \cite{dropout}.

We compare the performance of our proposed method with other methods for conditional density estimation on these univariate tasks. Specifically, we assess the log likelihood of the test set using our method against various established methods including NGBoost \citep{duan2020ngboost}, PGBM \citep{pgbm}, RoNGBa \citep{ren2019rongba}, TreeFlow \citep{wielopolski2023treeflow}, enhanced versions of KMN and MDN (``KMN+'' and ``MDN+'') \citep{rothfuss2019conditional}, Bayesian radial normalizing flows (RNF) \citep{bayesiannf}, Bayesian neural networks with homoscedastic Gaussian likelihoods using a mean-field variational approximation (MF), Mixture Density Networks (MDN) \citep{mdn}, neural networks with latent variable inputs (LV) \citep{lv}, Bayesian neural networks with homoscedastic Gaussian likelihoods using Hamiltonian Monte Carlo (HMC) \cite{bui2016deep}, and Bayesian neural networks with dropout (Dropout) \citep{dropout}. 
For our method, the algorithm initially fits tree-CDFs with node-level Logistic Regression and a maximum tree depth of 6. It then switches to node-level MLP with hidden layers sized (4,4) with the maximum tree depth reduced to 4. The hyperparameter that controls the level of $l_1$ penalty on the imbalanced splits in our method, $\eta$ (detailed in Appendix~\ref{a:single_tree}), is set to 0.1. 
For scale-specific shrinkage, we set \(c_0= 0.05\) and \(\gamma= 0.5\). The detailed experimental specifications and source of results are available in the Appendix~\ref{a-exp}. 
As an ablation experiment of the scale-specific shrinkage rate, we set $\gamma=0$ and held all other hyperparameter values the same as the current setting, and  
The detailed experimental specifications and source of results are available in the Appendix~\ref{a-exp}. 

\begin{figure}[h]
    \centering
    \includegraphics[width=\linewidth]{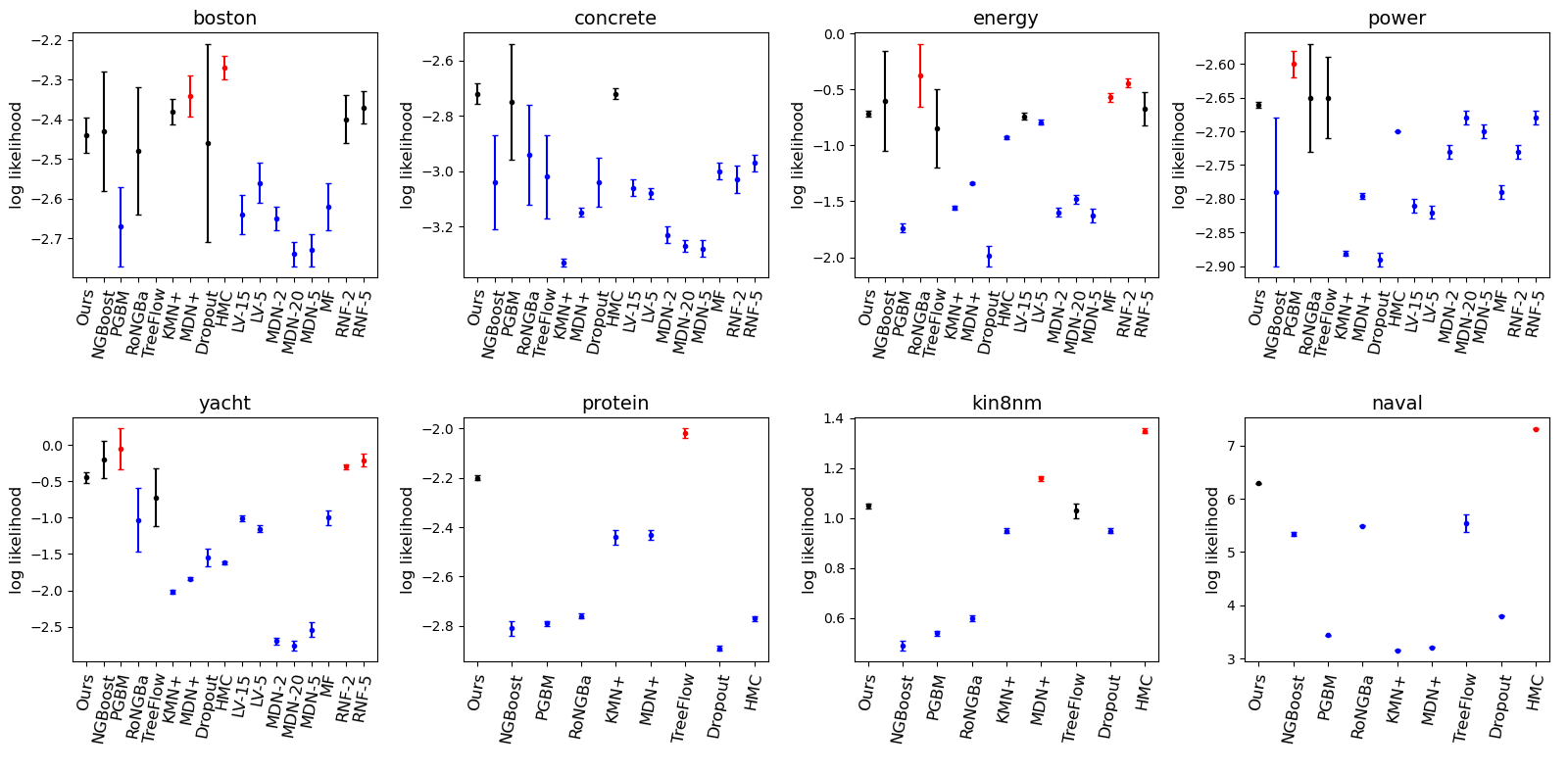}
    \caption{Comparison on UCI benchmark datasets as measured by log-likelihood of test set (mean $\pm$ standard error). Marker color indicates relative performance: {\color{blue}blue} indicates our method outperforms the alternative method, while {\color{red}red} indicates the instances when our method underperforms, and black denotes comparable performance within the standard error bounds. The results of NGBoost \citep{duan2020ngboost}, RoNGBa\citep{ren2019rongba}, and TreeFlow\citep{wielopolski2023treeflow} are obtained from their original papers. The results of PGBM\citep{pgbm} are obtained from \cite{wielopolski2023treeflow}. The results of Dropout, LV, MDN, MF, RNF are obtained from \cite{bayesiannf}. }
    \label{fig:uniy}
\end{figure}

The results are shown in Figure~\ref{fig:uniy}. Our method achieved the highest log likelihood on two datasets and outperformed most other methods on the remaining datasets. No other method consistently outperforms ours. Furthermore, the standard errors associated with our method are competitively low.

Additional experiments with variants of our proposed method (Table~\ref{ablation1} in Appendix~\ref{a:ablation}) show that flexible splits dominates constrained splits in the middle, and in most cases, the combination of Logistic Regression and MLP outperforms than either of the two, indicating different classifiers can indeed extract different aspects of the conditional distributions from the training data. 

To quantify the impact of scale-specific learning rates, we conducted an ablation experiment by setting $\gamma=0$ (holding all other hyperparameters unchanged). As shown in table~\ref{ablation_gamma0} in Appendix~\ref{a: additional experimental results}, the proposed method with scale-specific learning rates (with $\gamma > 0$) outperforms that with a constant learning rate on most of the datasets.

\subsection{Simulation examples for bivariate outcomes}
\label{sec:2dexp}
We assess our method using some challenging tasks involving bivariate outcome, originally proposed in \cite{ddn}. The conditional densities are shown in Figure~\ref{fig:2d-dens}, and detailed settings are provided in Appendix~\ref{a-exp}. For each task, the training set consists of 2000 observations generated from the joint probability distributions \(p(x, y_1, y_2)\).

\begin{figure}
    \centering
    \includegraphics[width=0.9\linewidth]{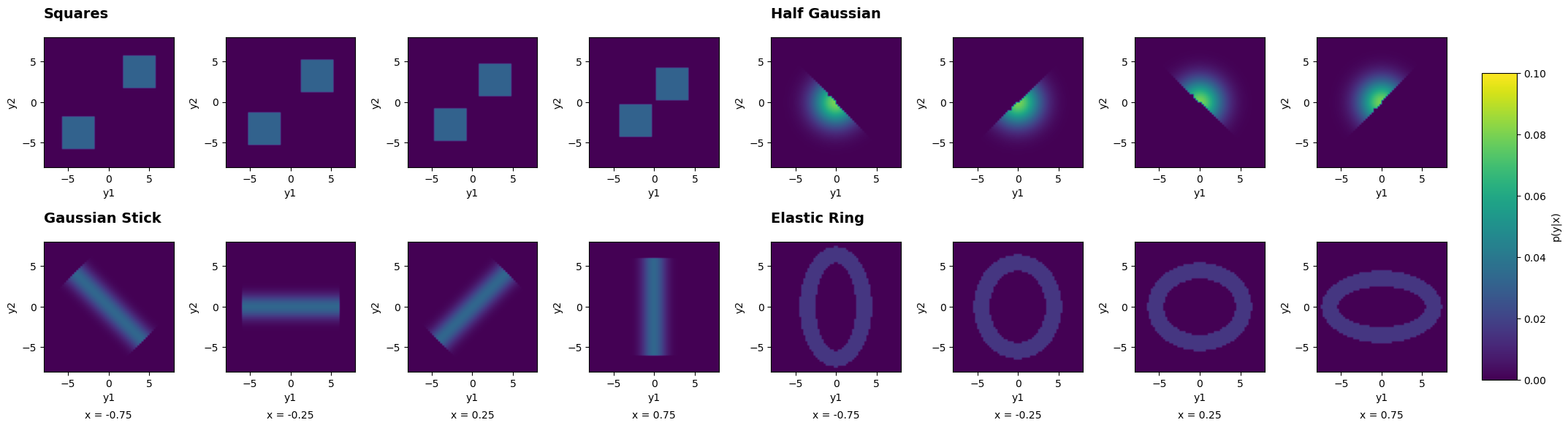}
    \caption{Ground truth conditional density of simulation examples with bivariate outcome}
    \label{fig:2d-dens}
\end{figure}

Our tree-flow is trained with the same hyperparameters and specifications as used in comparisons with other methods in Section~\ref{sec:1dexp}. Training on one simulated dataset with 2000 samples takes 405 seconds, 414 seconds, 407 seconds, and 580 seconds under the four scenarios respectively (using a single core on a MacBook Air equipped with an Apple M2 chip and 16GB RAM). 

For these simulated examples, the ground truth of the conditional density is analytically available, and the sum of squared errors (SSE) calculated on a $64\times 64$ grid of values of $(y_1,y_2)$ is used to measure the difference between the estimated conditional density and the ground truth. 
As shown in Table~\ref{tab:toy-sse}, our method achieves the lowest SSE under all scenarios. Applying rotations substantially reduces the SSE under all scenarios, while the performance without rotation is already competitive.
Visual comparisons between the ground truth and the conditional densities estimated by our method with and without rotations are included in Appendix~\ref{a:fig}. Incorporating rotations appears to help our method capture the non-smoothness of the conditional distribution with a boundary rotating with the value of $x$. An illustration of this effect can be seen in the half-Gaussian scenario presented in Figure~\ref{fig:halfgaussian} in Appendix~\ref{a:fig}.

\begin{table}[h]
\caption{SSE between ground truth and estimated conditional densities, averaged over four \(x\) values (-0.75, -0.25, 0.25, 0.75). The standard error of our method is calculated based on 20 runs. Lower SSE indicates better performance. The SSEs of the other methods being compared are obtained from \citep{ddn}, where the standard errors are not provided.}
\begin{center}

\begin{tabular}{lllll}
\hline
        ~ & Squares & Half-Gaussian & Gaussian Stick & Elastic Ring \\ 
        \hline
        DDN & 0.070 & 0.099 & 0.065 & 0.056 \\ 
        MAF & 0.224 & 0.088 & 0.106 & 0.172 \\ 
        MDN & 0.273 & 0.219 & 0.256 & 0.424 \\ 
        NSF & 0.149 & 0.173 & 0.077 & 0.235 \\ 
        RNF & 0.151 & 0.134 & 0.052 & 0.075 \\ 
        ours (no rotations) & 0.071$\pm$0.002 & 0.111$\pm$0.002& 0.061$\pm$0.002&0.057$\pm$0.001\\
        ours (12 rotations) &\textbf{0.055$\pm$0.001}& \textbf{0.071$\pm$0.001} & \textbf{0.041$\pm$0.001} & \textbf{0.044$\pm$0.001} \\ 
        \hline
    \end{tabular}
    \end{center}    
    \label{tab:toy-sse}
\end{table}

\subsection{Real-world tasks with multivariate outcomes}
\label{sec:ndexp}
We also evaluated our proposed method by comparing it with eight neural network approaches for conditional density estimation on UCI benchmark datasets involving multivariate outcomes. We did not include the gradient boosting methods evaluated in the univariate tasks (NGBoost, RoNGBa, and PGBM) in this comparison because their software implementations are designed only for univariate outcomes.
The characteristics of the UCI datasets with multivariate outcomes are shown in Table~\ref{tab:mvy_uci} in Appendix~\ref{a: additional experimental results}. Following \cite{ddn}, in each trial, each dataset is split into a training set and a test set with a ratio of 3:7 to create a data deficiency scenario, and both the covariates and outcomes are standardized using z-score normalization.

The average log-likelihood of the test set is compared in Table~\ref{tab:ll_uci_mvy}. For our method, 12 equi-spaced rotations are applied to datasets with 2-dimensional \(y\), and for ``air'' and ``skillcraft,'' 6 equi-spaced rotations are applied to each pair of coordinates of \(y\). \(\mathcal{X}\) is partitioned into 8 bins to average the rotations. (Based on our observations, the results are to some extent robust to the way of partitioning \(\mathcal{X}\). See Table~\ref{tab:nxbin} in Appendix~\ref{a: additional experimental results} for an example.) Table~\ref{tab:ll_uci_mvy} shows that our method achieves competitive performance. 
The results further demonstrate that rotations help our method adapt to real-world multivariate distributions, even when it is unknown whether there is an intrinsic rotation determined by \(\mathcal{X}\). The ensemble of rotations not only improves the average log-likelihood of our method but also enhances the stability and reduces the standard error of the estimated densities in both real-world tasks and the simulation examples. Similar to the experiments with univariate outcomes, our method achieves competitively low standard errors among the methods compared. (Full details on the datasets and the hyperparameter settings are available in Appendix~\ref{a-exp}.)

A comparison of the results obtained from different \(c_0\) and \(\gamma\) values is provided in Table~\ref{tab:c0gamma} in Appendix~\ref{a: additional experimental results}. The results align with our expectation that a smaller \(c_0\) and scale-specific shrinkage with a reasonably large \(\gamma\), which impose stronger regularization, would enhance the performance of our proposed method under this data deficiency scenario.

\begin{table}[h]
\caption{Comparison of log-likelihood on real-world tasks (mean$\pm$standard error). Methods with the best results are in bold; multiple bold methods indicate no significant differences. MDN+ and KMN+ results were obtained by running the respective software. Results for MDN, MAF, NSF, RNF, MLP, and DDN are from \citep{ddn}; ``NA" indicates results not provided in \citep{ddn}, and ``-Inf" indicates $-\infty$ log-likelihood in multiple runs. $\eta$ is a tuning parameter that controls the $l_1$ penalty on imbalanced splits in our method, detailed in Appendix~\ref{a:single_tree}.}
\begin{center}
    \begin{tabular}{llllll}
    \hline
        &Energy & Parkinsons & Temperature & Air & Skillcraft \\ \hline
       {MDN+} &1.33$\pm$0.02&-0.97$\pm$0.01&\bf{-0.64$\pm$0.01}&-1.01$\pm$0.01&-Inf*\\
        {KMN+} &1.16$\pm$0.03&-0.60$\pm$0.01&-0.91$\pm$0.01&-1.65$\pm$0.01&-Inf*\\
        MDN & -8.28$\pm$0.91 & -3.82$\pm$0.08 & -4.24$\pm$0.04 & -2.16$\pm$0.06 & -8.54$\pm$0.14 \\ 
        MAF & -125$\pm$51 & -20.1$\pm$1.7 & -14.0$\pm$0.4 & -14.5$\pm$1.4 & -81.1$\pm$8.2 \\ 
        NSF & -2.87$\pm$0.11 & -1.81$\pm$0.03 & -2.95$\pm$0.04 & 0.47$\pm$0.11 & -8.68$\pm$0.09 \\ 
        RNF & -19.4$\pm$4.2 & -4.01$\pm$0.25 & -7.51$\pm$0.62 & -0.81$\pm$0.26 & -26.8$\pm$2.2 \\ 
        MLP & -3.48$\pm$0.04 & -4.86$\pm$0.06 & -14.01$\pm$0.04 & NA & NA \\ 
        DDN  & 0.14$\pm$0.32 &\textbf{ -0.14$\pm$0.01} & -0.71$\pm$0.02 & 1.22$\pm$0.02 & \textbf{-1.56$\pm$0.02} \\
        DDN (no VL) & -1.56$\pm$0.27 & -0.17$\pm$0.02 & -0.84$\pm$0.02 & \textbf{1.32$\pm$0.02} & \textbf{-1.59$\pm$0.03} \\ 
        \hline
        ours ($\eta=0.1$)
        & \textbf{{1.84$\pm$0.04}} & -0.54$\pm$0.01 & {-0.68$\pm$0.01} & -0.67$\pm$0.01 & -1.66$\pm$0.02 \\ 
        ours ($\eta=0.01$) 
        & \textbf{1.86$\pm$0.04} & -0.56$\pm$0.01  & {-0.72$\pm$0.01} & -0.62$\pm$0.01 & \textbf{-1.57$\pm$0.02} \\ 
        ours ($\eta=0.1$, no rot.) 
        & 1.45$\pm$0.07 & -0.77$\pm$0.01 & -0.82$\pm$0.01 & -0.83$\pm$0.01 & -1.95$\pm$0.02 \\ 
      
        \hline
    \end{tabular}
\end{center}
    
    \label{tab:ll_uci_mvy}
\end{table}

\begin{figure}[!h]
    \centering
    \includegraphics[width=0.45\textwidth]{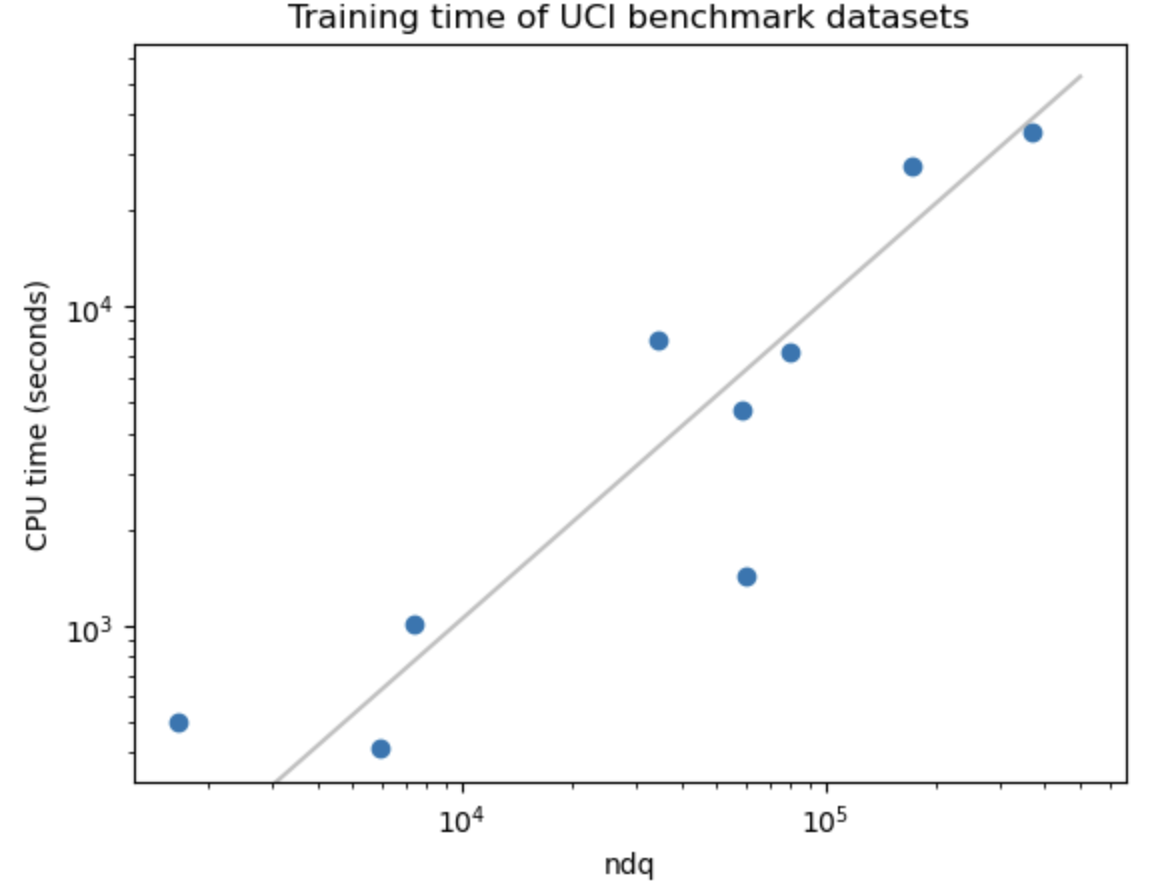}
    \caption{Training time of our method on a single CPU core versus $ndq$ on log-log scale for 9 UCI datasets—boston, concrete, power, yacht, naval, kin8nm, protein, air, and skillcraft. Points are annotated with $(n,d,q)$ values. A linear trend with slope~1 (gray line) indicates $O(ndq)$ complexity.}
    \label{fig1}
\end{figure}

The linear time complexity for training the tree flow is empirically confirmed across 9 UCI benchmark datasets, as shown in Figure~\ref{fig1}. Deviations from the linear trend are due to the varying number of trees required for each dataset. Based on our observations, tens to hundreds of trees are sufficient for the datasets used in this paper.
The training process depicted in Figure~\ref{fig1} does not utilize parallelization, and further improvement is expected because training the binary classifiers on the nodes within the same level of a tree can be completed in parallel. 

\subsection{Data generation}
\label{sec:diab}
We demonstrate the data generation capabilities of our proposed model using microbiome compositional data from 16S sequencing experiments. The DIABIMMUNE dataset \citep{diabimmune} includes microbiome compositions from 777 stool samples collected from 33 infants over a period of three years. For each observation \((x_i,y_i)\), the covariate \(x_i\) is the age at collection, and the outcome \(y_i\) is the microbiome composition at the operational taxonomic unit (OTU) level. The outcomes are normalized to relative abundances, i.e., the elements of each \(y_i\) sum to 1. We keep the 100 OTUs with the highest relative abundance. The proposed model was trained on the full dataset with \(c_0 = 0.1\), \(\gamma=0.5\), \(\eta=0.1\), and maximum depth of the trees is set to 4. No rotations were applied. With the trained model, one sample is generated for each $x_i$, mimicking the conditions under which the original data were collected.

\begin{figure}[h]
    \centering
    \includegraphics[width=\linewidth]{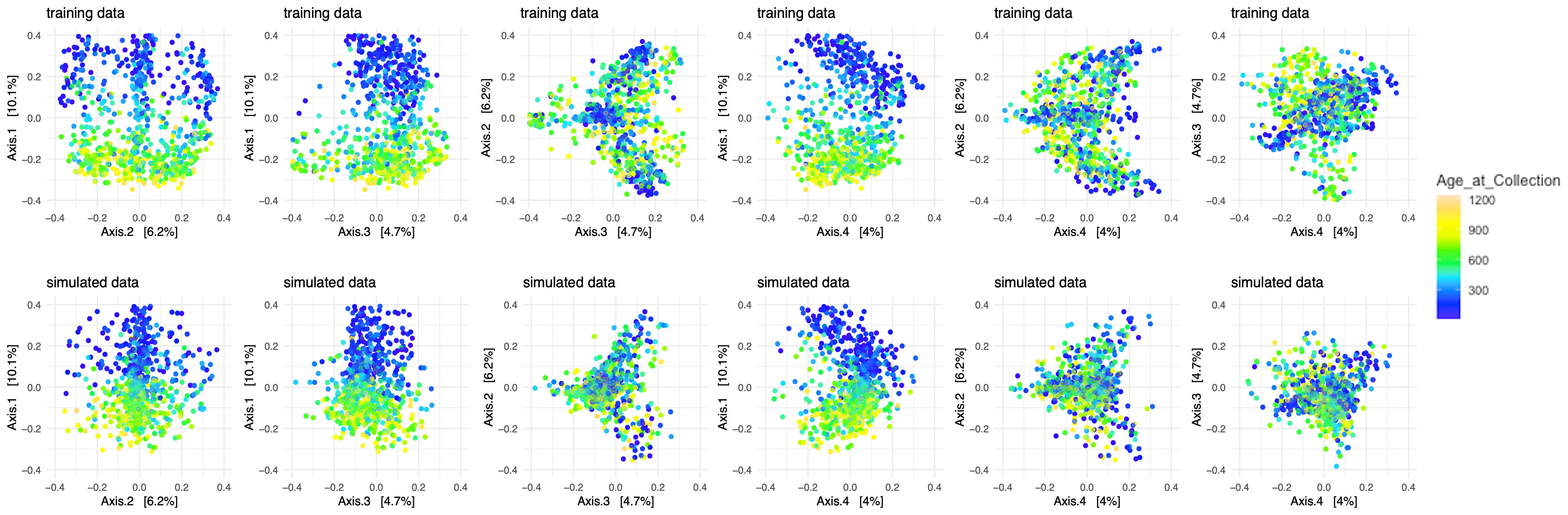}
    \caption{Principal coordinate analysis (PCoA) of Bray-Curtis similarity of training (upper row) and simulated (lower row) samples. The color of the points indicates the age (in days) of the infant, which is the covariate in this example.} 
    \label{fig:pcoa}
\end{figure}

Figure \ref{fig:pcoa} displays a principal coordinate analysis (PCoA) of the Bray-Curtis similarity of training and simulated samples. The simulated samples show similar marginal and conditional distributions to the training data, particularly in the lower-dimensional subspaces defined by the first four main axes of the PCoA.

\section{Conclusion}
\label{sec:limitation}
We proposed a generative model for conditional densities based on a normalizing flow with tree-CDF transforms. We demonstrated conditional density estimation with our proposed model and compared with other conditional density estimation methods with simulated data and real-world UCI datasets. We note that the performance of our method in the experiments is achieved with Logistic Regression and MLP(4,4), and expect a wider class of classifiers may provide further improvements. We also demonstrated the use of the proposed method in generative sampling in a microbiome context. Among many possible applications, the trained generative model can be used to provide uncertainty quantification on summary statistics computed on the microbiome data given the covariates.

A limitation of our approach, which is common for tree-based approaches adopting axis-aligned partitions, is that it may not approximate well high-dimensional distributions (i.e., those with hundreds or more features) especially in the presence of high-order correlation structure. So far our experiments have focused on tabular data with $<100$ dimensions, and so the available empirical evidence is limited to this domain. Possible extensions to overcome high-dimensional problems include adopting non-axis-aligned partitions, which will incur computational challenges. We leave this to future work.  

\section*{Acknowledgment}
This research is partly supported by NSF grant DMS-1749789 and NIGMS grant R01-GM135440.

\bibliographystyle{chicago}
\bibliography{arxiv}

\newpage 

\appendix
\section*{Appendix}
\section{Tree-CDF and its inverse}
\label{a:treecdf}
The multi-scale decomposition of tree-CDFs and their inverses, along with their properties, are detailed in \cite{awaya2023}. Here, we summarise these in the context of conditional density estimation. 

Suppose a probability measure \( G_x \) on \((0,1]^d\) is defined by a binary tree \( T \) with splitting probabilities \( G_x(A_l|A) = c(A)p_{\theta_{A}}(x) + (1-c(A))(\frac{\mu(A_l)}{\mu(A)}) \) for \( A \in I(T) \), and its corresponding tree-CDF is \( {\bf G}_x \). Then, for a \( d \)-dimensional vector \( y \) within \((0,1]^d\), where the path from the leaf containing \( y \) to the root is represented as \( y \in A_R \subset A_{R-1} \subset \cdots \subset A_1 = (0,1]^d \), applying \( {\bf G}_x \) to \( y \) involves a sequence of linear transforms along this path:
\[
{\bf G}_x(y) = {\bf G}_{x,A_1} \circ \cdots \circ {\bf G}_{x,A_{R-1}}(y),
\]
where each \( {\bf G}_{x,A} \) is defined based on the probability assignments at node $A$. For a node \( A = (a_{1},b_{1}] \times \cdots \times (a_{d},b_{d}] \) split along the \( j \)-th axis into \( A_l = (a_{1},b_{1}] \times \cdots \times (a_{j},s_{j}] \times \cdots \times (a_{d},b_{d}] \) and \( A_r = (a_{1},b_{1}] \times \cdots \times (s_{j},b_j] \times \cdots \times (a_{d},b_{d}] \), the transformation \( {\bf G}_{x,A} \) is given by:
\[
{\bf G}_{x,A}(y)[j'] = y_{j'} \quad \text{for } j' \neq j,
\]
\[
{\bf G}_{x,A}(y)[j] = \frac{G_x(A_l|A)}{(s_j-a_j)/(b_j-a_j)} y_j + \left(1 - \frac{G_x(A_l|A)}{(s_j-a_j)/(b_j-a_j)}\right) b_j \quad \text{for } y \in A_l,
\]
\[
{\bf G}_{x,A}(y)[j] = \frac{1-G_x(A_l|A)}{(b_j-s_j)/(b_j-a_j)} y_j + \left(1 - \frac{1-G_x(A_l|A)}{(b_j-s_j)/(b_j-a_j)}\right) b_j \quad \text{for } y \in A_r.
\]

Let \(z_j = \frac{y_j-a_j}{b_j-a_j}\). The inverse node-level transform, ${\bf G}_{x,A}^{-1}(y)$, is given by 
\[\hat {\bf G}_{x,A}^{-1}(y)[j'] = y_{j'} \quad \text{ for } j'\neq j,\]
\[\hat {\bf G}_{x,A}^{-1}(y)[j] = a_j + \frac{c_j-a_j}{G_x(A_l|A)}z_j  \quad\text{if } z_j\leq G_x(A_l|A),\]
\[\hat {\bf G}_{x,A}^{-1}(y)[j] = c_j + \frac{b_j-c_j}{1-G_x(A_l|A)}(z_j-G_x(A_l|A))  \quad\text{if } z_j\leq G_x(A_l|A).\]

It can be seen from the above formula that the time complexity of applying \( {\bf G}_x \) or \( {\bf G}_x^{-1} \) is equivalent to that of calculating \( p_{\theta_{A}}(x) \) if the maximum depth of the trees is fixed.

\section{Justification of optimization procedure in \ref{sec:single_tree}}
\label{a:single_tree}

For an observation $(x_i,y_i)$ where $y_i$ belongs to a leaf node $L$, we have 
$g_{x_i}(y_i)=\frac{G_{x_i}(L)}{\mu(L)}.$

$G_{x_i}(L)$ and $\mu(L)$ can be decomposed on the tree as a product of splitting probabilities:
$$G_{x_i}(L) = \prod_{A\in I(T),y_i\in A}G_{x_i}(A_l|A)^{\mathbf{1}(y_i\in A_l)}G_{x_i}(A_r|A)^{\mathbf{1}(y_i\in A_r)},$$
and
$$\mu(L) = \prod_{A\in I(T),y_i\in A}(\frac{\mu(A_l)}{\mu(A)})^{\mathbf{1}(y_i\in A_l)}(\frac{\mu(A_r)}{\mu(A)})^{\mathbf{1}(y_i\in A_r)}.$$
Therefore, 
\small{
\begin{align*}
    l(T,\theta) =&\sum_i\log g_{x_i}(y_i)\\
    =&\sum_i\sum_{y_i\in A \in I(T)}{\mathbf{1}(y_i\in A_l)}\log G_{x_i}(A_l|A) + \mathbf{1}(y_i\in A_r)\log G_{x_i}(A_r|A) - {\mathbf{1}(y_i\in A_l)}\log\frac{\mu(A_l)}{\mu(A)}- \mathbf{1}(y_i\in A_r)\log\frac{\mu(A_r)}{\mu(A)}\\
    =&\sum_{A\in I(T)}\sum_{i:y_i\in A}\left({\mathbf{1}(y_i\in A_l)}\log G_{x_i}(A_l|A) + \mathbf{1}(y_i\in A_r)\log G_{x_i}(A_r|A)\right) - n(A_l)\log\frac{\mu(A_l)}{\mu(A)}- n(A_r)\log\frac{\mu(A_r)}{\mu(A)}\\
    =&\sum_{A\in I(T)}\left(\sum_{i:y_i\in A}{\mathbf{1}(y_i\in A_l)}\log p_{\theta_A}(x_i) + {\mathbf{1}(y_i\in A_r)}\log (1-p_{\theta_A}(x_i)) \right) + \left(- n(A_l)\log\frac{\mu(A_l)}{\mu(A)}- n(A_r)\log\frac{\mu(A_r)}{\mu(A)}\right)\\
    =&\sum_{A\in I(T)} \left(l_{A,{
    \rm bin}}(T,\theta_A) + C_A(T) \right).
\end{align*}
}
This proves Eq~\ref{eq:ll} in Section~\ref{sec:single_tree}.

We use a greedy, root-to-leaf tree learning algorithm, where the tree is expanded by splitting one node at a time based on maximizing $l(T,\theta)$ after the current splitting. 
Following the notations in Section~\ref{sec:single_tree}, the tree is initialized as $T_0^*$ with only the root node, and $\theta_0^*$ is an empty set. Suppose the current tree and corresponding node-level parameters are $(T_j^*, \theta_j^*)$, then $T_j^*$ is chosen among the $M$ candidates $T_{j,1},\cdots,T_{j,M}$ to maximize $l(T,\theta)$:
$$(T_j^*,\theta_j^*) = \argmax_{T\in\{T_{j,1},\cdots, T_{j,M}\},\theta} l(T,\theta).$$
Since $T_{j,1},\cdots,T_{j,M}$ only differ by the way of splitting $A$, with the decomposition of the log-likelihood shown above, we have
$$\underset{T \in\left\{T_{j, 1}, \cdots, T_{j, M}\right\}, \theta}{\arg \max } l(T, \theta) = \argmax_{T\in\{T_{j,1},\cdots, T_{j,M}\},\theta}l_{A,{\rm bin}}(T,\theta_A) + C_A(T). $$

Given the tree structure $T$, 
since $C_A(T)$ does not involve $\theta$, we have 
$$\argmax_\theta( l_{A,{\rm bin}}(T,\theta_A) + C_A(T)) = \argmax_\theta l_{A,{\rm bin}}(T,\theta_A) \quad \text{for any } T.$$
Let $\theta_A^*(T)=\argmax_\theta l_{A,{\rm bin}}(T,\theta_A)$. We have
\begin{align*}
&\max_{T \in\left\{T_{j, 1}, \cdots, T_{j, M}\right\},\theta}l_{A,{\rm bin}}(T,\theta_A)+C_A(T) = \max_{T \in\left\{T_{j, 1}, \cdots, T_{j, M}\right\}} (\max_\theta l_{A,{\rm bin}}(T,\theta_A)+C_A(T)) \\
&= \max_{T \in\left\{T_{j, 1}, \cdots, T_{j, M}\right\}} l_{A,{\rm bin}}(T,\theta_A^*(T)) + C_A(T),    
\end{align*}
therefore 
$$
T_j^*=\underset{T \in\left\{T_{j, 1}, \cdots, T_{j, M}\right\}}{\arg \max } l_{A,{\rm bin}}\left(T, \theta_A^*(T)\right) + C_A(T),
$$
and
$$
\theta^*_A=\theta_A^*(T_j^*).
$$
This justifies the two-step training algorithm described in Section~\ref{sec:single_tree}.

In practice, one can incorporate further penalty terms on the complexity of tree into $C_A(T)$ without affecting the decomposition of $l(T,\theta)$. In our implementation, we used an $l_1$ penalty on imbalanced splits. Specifically, if $A$ is split along the j-th axis at $s_j$, and $A = (a_1,b_1]\times \cdots \times (a_j,b_j] \times \cdots \times (a_d, b_d]$,  then an $l_1$ penalty term on imbalanced split is defined as 
$$L_\eta(s_j) = -\eta|s_j-(a_j+b_j)/2|,$$ 
where $n(A) = \sum_{i=1}^n 1(y_i^{(k)}\in A)$ is the number of samples within node $A$, $\eta$ is a hyperparameter. With such penalty term, $C_A(T)$ becomes
$$C_A(T) = - n(A_l)\log\frac{\mu(A_l)}{\mu(A)}- n(A_r)\log\frac{\mu(A_r)}{\mu(A)} + L_\eta(s_j)$$
if the node $A$ of $T$ is split at $s_j$.

\section{Algorithms for training the tree flow and a single tree}
\label{a:algo}
The algorithm for training the tree flow $\mathbf{G}_x$ is given in Algorithm~\ref{alg:ensemble}. 

\begin{algorithm}[h]
   \caption{Training the tree flow}
   \label{alg:ensemble}
\begin{algorithmic}[1]
   \State {\bfseries input:} training data \(\{(x_i, y_i)\}_{i=1}^n\), validation data \(\{(x_{i,val}, y_{i,val})\}_{i=1}^n\), maximum number of trees \( K_{max} \), window size \( w \), shrinkage parameters (\( c_0, \gamma \))
   \State {\bfseries Output:} \( \{\mathbf{G}_{k,x}\}_{k=1}^K \)
   \State Initialize \( \text{LL}^{(0)} \gets 0 \)
   \For{\( i = 1 \) \textbf{to} \( n \)}
       \State \( y_i^{(0)} \gets y_i \) 
        \State \( y_{i,val}^{(0)} \gets y_{i,val} \)
   \EndFor
   \For{\( k = 1 \) \textbf{to} \( K_{max} \)}
       \State Train \( G_{k,x} \) on \(\{(x_i, y_i^{(k-1)})\}\) using Algorithm~\ref{algo:single_tree}
     
       \State Shrink \(G_{k,x}\)   \Comment{As described in Section~\ref{sec:additional}}
      
       \State \( \mathbf{G}_{k,x} \) $\leftarrow$ tree-CDF of \( G_{k, x} \)
     
       \State \( \text{LL}^{(k)} \leftarrow \text{LL}^{(k-1)} + \sum_{i=1}^n \log g_{k, x_i}(y_{i,val}^{(k-1)}) \) \Comment{Update log-likelihood}
       \For{\( i = 1 \) \textbf{to} \( n \)}
       \State \( y_i^{(k)} \gets \mathbf{G}_{k,x_i}(y_i^{(k-1)}) \) 
       \State \( y_{i,val}^{(k)} \gets \mathbf{G}_{k,x_{i,val}}(y_{i,val}^{(k-1)}) \) 
       \Comment{Update observations}
   \EndFor
       \If{\( \text{LL}^{(k)} - \text{LL}^{(k-w)} \leq 0 \)} \Comment{Early stopping}
           \State \textbf{break}
       \EndIf
   \EndFor
\end{algorithmic}
\end{algorithm}

Algorithm~\ref{algo:single_tree} summarizes the algorithm for fitting a single tree-CDF.

\begin{algorithm}
\caption{Training a single tree-CDF}
\label{algo:single_tree}
\begin{algorithmic}[1]
\Procedure{TrainTreeCDF}{maxDepth, min\_samples, $\{x_i,y_i\}$, root,treeCandidates}
    \State $\text{queue} \gets$ [root] 
    \State $\text{currentDepth} \gets 0$
    \While{$\text{currentDepth} < \text{maxDepth}$}
        \State $\text{levelSize} \gets$ len(queue)
        \For{$i \gets 1$ to $\text{levelSize}$}
            \State $A \gets \text{queue.pop}(0)$
            \If{n(A)$<$min\_samples}
            \State continue
            \EndIf
            \State $T\gets$optimalSplit(A,treeCandidates)
            \Comment{Update $A_l, A_r$ accordingly}
            \State $\theta_A \gets \text{fitBinaryClassification}(T,A,\{x_i,y_i\})$ 
            \State queue.append$(A_l)$
            \State queue.append$(A_r)$
        \EndFor
        \State $\text{currentDepth} \gets \text{currentDepth} + 1$
    \EndWhile
    \State \Return root
\EndProcedure

\State \textbf{function} fitBinaryClassification(T, A,$\{x_i,y_i\}$)    
    \Return $\arg\max_{\theta_A} l_{A,{\rm bin}(T,\theta_A)}$\Comment{Defined in Section\ref{sec:single_tree}}
\State \textbf{end function}

\State \textbf{function} optimalSplit(A,treeCandidates,$\{x_i,y_i\}$)
    \State $LL_{\text{max}} \gets -\infty, T^*\gets$ None
    \For{$T$ in treeCandidates[A]}
        \State $\theta_A \gets \text{fitBinaryClassification}$(T,A, $\{x_i,y_i\}$)
       \State $LL\gets l_{A,{\rm bin}}(T, \theta_A) + C_A(T)$ \Comment{Defined in Section\ref{sec:single_tree}}
        \If{$LL > LL_{\text{max}}$}
            \State $LL_{\text{max}} \gets LL$ 
            \State $T^*\gets T$
        \EndIf
    \EndFor
\State \Return $T^*$
\State \textbf{end function}
\end{algorithmic}
\end{algorithm}

\section{Time complexity analysis}
\label{a:time_complexity}
To fit a tree-CDF, the optimal splitting at each internal node is selected from \( S \times d \) possible splits (with \( S \) cutpoints per axis). When evaluating each candidate split, the fitting process for a node-level binary classifier—whether using Logistic Regression or a Multilayer Perceptron with two hidden layers of 4 nodes each as in our implementation—has a time complexity of \( O(nq) \). Therefore, the overall complexity for fitting each tree-CDF is \( O(ndq) \).

Applying a tree-CDF to one observation indeed has a complexity of \( O(q) \). As demonstrated in previous work \citep{awaya2023}, a tree-CDF can be represented by a series of linear transformations at each level of the tree, with each transformation costing \( O(q) \) due to the evaluation of the splitting probability with the trained node-level binary classifier. Since the maximum depth \( R \) of the trees is fixed, there are at most \( R \) of these \( O(q) \) operations, thus applying a tree-CDF to a $d$-dimensional vector is \( O(q) \). (Detailed information about the multi-scale decomposition of tree-CDFs and their inverses are provided in the Appendix~\ref{a:treecdf}. )

With the fitted conditional tree flow, evaluating the density of a test sample is \( O(q) \) because it avoids any computationally expensive steps such as evaluating Jacobians. Instead, the following equation is used for density evaluation of a sample \( (x, y) \):
\[
f_x(y) = \prod_{k=1}^K g_{k, x}(y^{(k-1)})
\]
where \( y^{(k)} = G_{k, x}(y^{(k-1)}) \) and \( y^{(0)} = y \). Updating \( y^{(k-1)} \) to \( y^{(k)} \) is \( O(q) \), and calculating \( g_{k, x}(y^{(k-1)}) \) involves just the product of splitting probabilities along the path from the root to the leaf that contains \( y^{(k-1)} \) divided by the volume of the leaf, which is also at most \( O(q) \). 

Sampling from the fitted conditional tree flow given \( x \) involves applying the inverse tree-CDFs, \( \mathbf{G}_{K, x}^{-1}, \cdots, \mathbf{G}_{1, x}^{-1} \), to a uniform random variable. The inverse tree-CDF employs a similar multi-scale decomposition as the tree-CDF, and applying an inverse tree-CDF to a \( d \)-dimensional vector is \( O(q) \) due to the \( O(q) \) time complexity of evaluating the splitting probabilities given \( x \). Therefore, drawing one sample also has a time complexity of \( O(q) \).

\section{Rotation ensemble of tree flows}
\label{a:rotation}
Suppose the $y_i$'s in the original training data are rotated to generate \(J\) distinct data sets, denoted as \(D_1, \cdots, D_J\), where 
\(D_j = \{(x_i, y_iR_j)\}\), \(R_j\) is a rotation matrix applied to each data set. Training the conditional tree flow on $D_j$ yields $f_x^{(j)}$. Note that rotations are orthogonal transformations, the resulting conditional density is defined as
\begin{equation}
f_x(y) = \sum_{j=1}^J w_x^{(j)} f_x^{(j)}(yR_j),
\end{equation}

The weights \( w_x^{(j)} \) are dependent on \( x \) and are calculated based on the partitioning of the feature space \( \mathcal{X} \) into disjoint regions \( X_1, \cdots, X_{K'} \). We assume that within each region \( X_k \), the weights remain constant for all points \( x \). Thus, for any \( x \in X_k \), the weight is computed by
\[
w_{x}^{(j)} = \frac{\prod_{x_i \in X_k} f_{x_i}^{(j)}(y_iR_j)}{\sum_{j'=1}^J\prod_{x_i \in X_k} f_{x_i}^{(j')}(y_iR_{j'})}.
\]

\section{Full experimental details}
\label{a-exp}
\subsection{Experiment settings and details} 
\textbf{Data dimensions}. Dimensions of the UCI datasets used in Sectiokn~\ref{sec:exp} are shown in Table~\ref{uci1} and Table~\ref{tab:mvy_uci}.

\begin{table}[h]
\caption{Characteristics of UCI datasets with univariate outcome}
\begin{center}

    \begin{tabular}{ccc }
    \hline
         Dataset & $n$ & $q$\\
         \hline
         boston & 506 & 13 \\ 
        concrete & 1030 & 8 \\ 
        energy & 768 & 8 \\ 
        power & 9568 & 4 \\ 
        yacht & 308 & 6 \\ 
        kin8nm & 8192 & 8 \\ 
        naval & 11934 & 17 \\ 
        protein & 45730 & 9 \\ 
        \hline
    \end{tabular}
    \end{center}
    
    \label{uci1}
\end{table}

\begin{table}[H]
    \caption{Characteristics of UCI datasets with multivariate outcome}
\begin{center}

\begin{tabular}{lccc}
\hline & $n$ & $q$ & $d$ \\
\hline 
Energy & 768 & 8 & 2 \\
Parkinsons & 5875 & 16 & 2 \\
Temperature & 7588 & 21 & 2 \\
Air & 8891 & 10 & 3 \\
Skillcraft & 3338 & 15 & 4 \\
\hline
\end{tabular}
    \label{tab:mvy_uci}
    
\end{center}
\end{table}

\textbf{Data splits}. For UCI datasets with univariate $y$, we use the train test splits provided by \\ \hyperref[]{https://github.com/yaringal/DropoutUncertaintyExps}. 
For UCI datasets with multivariate $y$, we use the same train test splits as \cite{ddn}. 
For simulation examples, training set and test set are generated independently from the ground truth. 
In each run, a random subset of the training data (comprising 10\% of the training data) is used as the validation set to determine early stopping. 

\textbf{Simulation settings (bivariate outcome)}.
Four conditional distributions of $y_1, y_2|x$ are considered:

\textit{Squares}: $x\sim U(-1,1)$, $\lambda \sim \operatorname{Bern}(0.5) , $
$
a_1, a_2 \stackrel{i i d}{\sim} U(x-5,x-1) , b_1, b_2 \stackrel{i i d}{\sim} U(1-x, 5-x) $,
$$
 y_1=\lambda a_1+(1-\lambda) b_1 , y_2=\lambda a_2+(1-\lambda)  b_2.
$$

\textit{Half Gaussian}: $x\sim U(-1,1)$, $a, b \stackrel{\text { iid }}{\sim} N(0,2)$, $$y_1=|a| \cos x \pi-b \sin x \pi , y_2=|a| \sin x \pi+b \cos x \pi.$$

\textit{Gaussian Stick}: $x\sim U(-1,1)$, $a \sim N(0,1) , b \sim U(-6,6)$, $c=(-0.75+x) / 2$,
$$y_1=a \cos c \pi-b \sin c \pi ,
y_2=a \sin c \pi+b \cos c \pi.
$$

\textit{Elastic Ring}: $x\sim U(-1,1)$, $d \sim U(0,2) , \theta \sim U(0,2 \pi)$,
$$
y_1=(4+2  x+d) \cos \theta , y_2=(4-2 x+d) \sin \theta.
$$

\textbf{Hyperparameters}. 
For our method, the choice of $c_0, \gamma$ is based on the recommendations in \citep{awaya2023}. We set $c_0 = 0.05, \gamma = 0.5$ for the experiments in the main paper, and the results obtained with other values of $c_0, \gamma$ is included in Appendix~\ref{a-c0gamma}. The splitting point is obtained by grid search over 20 equally-spaced gridpoints per axis. Maximum number of trees for training the flow with each type of binary classifiers is set to 1000 as an upper limit. In our experiments, the resulting $K$ from early stopping ranges from tens to hundreds. The early stopping window is set to 10. The minimum number of samples per node is set to 10. We observed in the experiments that the results are generally robust to the early stopping window and the minimum number of samples per node.    

For KMN+ and MDN+, We adopted the hyperparameter specifications \texttt{x\_noise\_std=0.2, y\_noise\_std=0.1} 
 as recommended in the experiments in \cite{rothfuss2019conditional}.

\textbf{Implementation details}. 
Within our model, the binary classifiers are the only components that need the use of optimization techniques for effective training. These classifiers are implemented using the sklearn library, with specific settings for each:
\begin{itemize}
    \item Logistic Regression: Fitted using \texttt{sklearn.linear\_model.LogisticRegression}, with the following configuration: \texttt{random\_state=42,max\_iter=1000, solver='lbfgs'}, and all other arguments are set to default. 
    \item Multilayer perceptron (MLP): Fitted using \texttt{sklearn.neural\_network.MLPClassifier}, with the following configuration: \texttt{random\_state=42,max\_iter=1000, solver='lbfgs', hidden\_layer\_sizes=(4,4)}, and all other arguments are set to default. 
\end{itemize}

\textbf{Source of experimental results}. 
For the univariate experiments, the results of NGBoost \citep{duan2020ngboost}, RoNGBa\citep{ren2019rongba}, and TreeFlow\citep{wielopolski2023treeflow} are obtained from their original papers. The results of PGBM\citep{pgbm} is obtained from \cite{wielopolski2023treeflow}. The results of Dropout, LV, MDN, MF, RNF are obtained from \cite{bayesiannf}. 
For the simulation examples and multivariate experiments, the results of MAF, MDN, NSF, RNF, MLP and DDN are obtained from \citep{ddn}.

\textbf{Source of existing code and datasets used in this work.}
The experiment results of KMN+ and MDN+ \cite{rothfuss2019conditional} are obtained using the code provided at \url{https://github.com/freelunchtheorem/Conditional_Density_Estimation}. 
The simulation examples with bivariate outcome are generated with code available at \url{https://github.com/NBICLAB/DDN}. 
For the UCI benchmark datasets, the original datasets are available at \url{https://archive.ics.uci.edu/}. We used the code at \url{https://github.com/yaringal/DropoutUncertaintyExps} to preprocess and split datasets for the experiments in Section~\ref{sec:1dexp}. Code for preprocessing and splitting datasets used in Section~\ref{sec:ndexp} is provided by the authors of \citep{ddn}.

\subsection{Experiments compute resources}

All experiments were conducted on a computing cluster where each experimental run utilized a single CPU; no experiments were performed using GPUs. The memory allocation for all runs was set to 2GB, which served as a generous upper limit and allowed for caching all intermediate results, although this was not necessary for producing the results presented in the paper.

The full research project did not require more compute than the experiments reported in the paper.

\section{Additional experimental results}
\label{a: additional experimental results}

\subsection{Effect of flexible splitting and combination of classifiers}
\label{a:ablation}
We aim to understand the contribution of flexible splitting and the combination of binary classifiers. We set $c_0 = 0.05, \gamma=0.5, \eta=0.1$, and set maximum depth of trees to 6, and 
assess the following variants of our methods:
(1) The \textit{full} model, where splits are obtained by grid search at each node, and node-level classification first uses Logistic Regression until early stopping criteria is met, then switches to MLP. (2) All nodes are constrained to be split in the \textit{middle}. Same as the full model, both LR and MLP are used. (3) Only use \textit{LR} at internal nodes. (4) Only use \textit{MLP} at internal nodes. 
\begin{table}[h]
    \caption{Average log likelihood (mean$\pm$standard error) of univariate tasks. Larger values indicate better performance.}

\begin{center}
    \begin{tabular}{lllll}
    \hline
        ~ & full & middle &LR & MLP \\ \hline
        boston & \textbf{-2.47$\pm$0.05} & \textbf{-2.53$\pm$0.05} & \textbf{-2.55$\pm$0.04} & -2.61$\pm$0.04 \\ 
        concrete & \textbf{-2.67$\pm$0.05 }& -2.77$\pm$0.04 & -3.47$\pm$0.02 & \textbf{-2.75$\pm$0.05} \\ 
        energy & \textbf{-0.75$\pm$0.04} & \textbf{-0.78$\pm$0.03} & -1.45$\pm$0.03 & -0.86$\pm$0.04 \\ 
        power & -2.69$\pm$0.01 & -2.73$\pm$0.01 & -2.84$\pm$0.01 & \textbf{-2.65$\pm$0.01} \\ 
        yacht & \textbf{-0.53$\pm$0.07} & -0.88$\pm$0.07 & -1.05$\pm$0.06 & -1.56$\pm$0.09 \\
        \hline
    \end{tabular}
    \label{ablation1}
    \end{center}
\end{table}

\subsection{Effect of scale-specific shrinkage rates}
The proposed method with scale-specific learning rates ($\gamma=0.5$) and fixed learning rates ($\gamma=0$) are compared in Table~\ref{ablation_gamma0}. The other hyperparameters are set to the same values as in Section~\ref{sec:1dexp}. 
 
\begin{table}[]
    \centering
\caption{Comparison of predictive scores for different datasets with $\gamma = 0$ and $\gamma =0.5$}
\begin{tabular}{lcc}
\toprule
\textbf{Dataset} & \textbf{$\gamma = 0$} & \textbf{$\gamma =0.5$} \\
\midrule
kin8nm                        & 0.99 $\pm$ 0.01 & 1.05 $\pm$ 0.01 \\
bostonHousing                 & -2.53 $\pm$ 0.05 & -2.44 $\pm$ 0.04 \\
power-plant                   & -2.68 $\pm$ 0.01 & -2.66 $\pm$ 0.01 \\
concrete                      & -2.78 $\pm$ 0.05 & -2.72 $\pm$ 0.04 \\
protein-tertiary-structure (dequantized) & -2.13 $\pm$ 0.01 & -2.13 $\pm$ 0.01 \\
yacht                         & -0.56 $\pm$ 0.10 & -0.45 $\pm$ 0.07 \\
naval-propulsion-plant        & 6.48 $\pm$ 0.01  & 6.29 $\pm$ 0.01 \\
\bottomrule
\end{tabular}
    \label{ablation_gamma0}
\end{table}

\subsection{Results shown in Figure~\ref{fig:uniy}}

\begin{table}[!h]
\caption{Comparison on UCI benchmark datasets, measured by the log-likelihood of the test set (mean $\pm$ standard error). 
Mean and standard error of the log-likelihood are calculated based on 20 runs, except for "protein", which is based on 5 runs. NA indicates that the results are not provided in the original paper.}
\label{tab1}
\scriptsize
    \centering
    \begin{tabular}{lllllllll}
    \hline
         & boston & concrete & energy & power & yacht & protein & kin8nm & naval \\ \hline
        Ours & -2.44$\pm$0.04 & {-2.72$\pm$0.04} & -0.72$\pm$0.03 & -2.66$\pm$0.01 & -0.45$\pm$0.07 & -2.20$\pm$0.01 & 1.05$\pm$0.01 & 6.29$\pm$0.01 \\ 
        NGBoost & -2.43$\pm$0.15 & {-3.04$\pm$0.17} & -0.60$\pm$0.45 & {-2.79$\pm$0.11} & -0.20$\pm$0.26 & {-2.81$\pm$0.03} & {0.49$\pm$0.02} & {5.34$\pm$0.04} \\ 
        PGBM & {-2.67$\pm$0.10} & -2.75$\pm$0.21 & {-1.74$\pm$0.04} & {-2.60$\pm$0.02} & {-0.05$\pm$0.28} & {-2.79$\pm$0.01} & {0.54$\pm$0.04} & {3.44$\pm$0.04} \\ 
        RoNGBa & -2.48$\pm$0.16 & {-2.94$\pm$0.18} & {-0.37$\pm$0.28} & -2.65$\pm$0.08 & {-1.03$\pm$0.44} & {-2.76$\pm$0.03} & {0.60$\pm$0.03} & {5.49$\pm$0.04} \\ 
        KMN+ & -2.38$\pm$0.03 & {-3.33$\pm$0.01} & {-1.56$\pm$0.02} & {-2.88$\pm$0.01} & {-2.02$\pm$0.03} & {-2.44$\pm$0.01} & {0.95$\pm$0.01} & {3.16$\pm$0.01} \\ 
        MDN+ & {-2.34$\pm$0.05} & {-3.15$\pm$0.02} & {-1.34$\pm$0.01} & {-2.80$\pm$0.01} & {-1.84$\pm$0.02} & {-2.43$\pm$0.01} & {1.16$\pm$0.01} & {3.21$\pm$0.01} \\ 
        TreeFlow & NA & {-3.02$\pm$0.15} & -0.85$\pm$0.35 & -2.65$\pm$0.06 & -0.72$\pm$0.40 & {-2.02$\pm$0.02} & 1.03$\pm$0.06 & {5.54$\pm$0.16} \\ 
        Dropout & -2.46$\pm$0.25 & -3.04$\pm$0.09 & -1.99$\pm$0.09 & -2.89$\pm$0.01 & -1.55$\pm$0.12 & {-2.89$\pm$0.01} & {0.95$\pm$0.01} & {3.80$\pm$0.01} \\ 
        HMC & {-2.27$\pm$0.03} & {-2.72$\pm$0.02} & {-0.93$\pm$0.01} & {-2.70$\pm$0.01} & {-1.62$\pm$0.02} & {-2.77$\pm$0.01} & {1.35$\pm$0.01} & {7.31$\pm$0.01} \\ 
        LV-15 & {-2.64$\pm$0.05} & -3.06$\pm$0.03 & -0.74$\pm$0.03 & -2.81$\pm$0.01 & -1.01$\pm$0.04 & NA & NA & NA \\ 
        LV-5 & {-2.56$\pm$0.05} & -3.08$\pm$0.02 & -0.79$\pm$0.02 & -2.82$\pm$0.01 & -1.15$\pm$0.05 & NA & NA & NA \\ 
        MDN-2 & -2.65$\pm$0.03 & -3.23$\pm$0.03 & -1.60$\pm$0.04 & -2.73$\pm$0.01 & -2.70$\pm$0.05 & NA & NA & NA \\ 
        MDN-20 & -2.74$\pm$0.03 & -3.27$\pm$0.02 & -1.48$\pm$0.04 & -2.68$\pm$0.01 & -2.76$\pm$0.07 & NA & NA & NA \\ 
        MDN-5 & -2.73$\pm$0.04 & -3.28$\pm$0.03 & -1.63$\pm$0.06 & -2.70$\pm$0.01 & -2.54$\pm$0.10 & NA & NA & NA \\ 
        MF & -2.62$\pm$0.06 & -3.00$\pm$0.03 & -0.57$\pm$0.04 & -2.79$\pm$0.01 & -1.00$\pm$0.10 & NA & NA & NA \\ 
        RNF-2 & -2.40$\pm$0.06 & -3.03$\pm$0.05 & -0.44$\pm$0.04 & -2.73$\pm$0.01 & -0.30$\pm$0.04 & NA & NA & NA \\ 
        RNF-5 & -2.37$\pm$0.04 & -2.97$\pm$0.03 & -0.67$\pm$0.15 & -2.68$\pm$0.01 & -0.21$\pm$0.09 & NA & NA & NA \\ 
        \hline
    \end{tabular}
\end{table}

\subsection{Additional results}
\label{a-c0gamma}
\ref{tab:nxbin} shows the average test log-likelihood of the UCI datasets with multivariate outcome with different number of bins for $X$ for rotations. The hyperparameters for our model are configured as follows: \(c_0 = 0.05\), \(\gamma = 0.5\), and \(\eta = 0.01\). The maximum depth of the trees, $R$, is set to 6 when using Logistic Regression and reduced to 4 when using Multilayer Perceptrons (MLP).
 The results are robust to the way of partitioning $\mathcal{X}$.  
\begin{table}[h]
\caption{Sensitivity analysis of partitions of $X$. }
\begin{center}

    \begin{tabular}{ccc}
    \hline
        data & partition of X & average test log-likelihood (mean$\pm$SE)\\ \hline
        energy & kmeans, k=4 & 1.865$\pm$0.043 \\ 
        energy & kmeans, k=8 & 1.863$\pm$0.042 \\ 
        energy & HDBSCAN & 1.864$\pm$0.043 \\ 
        parkinsons & kmeans, k=4 & -0.561$\pm$0.007 \\ 
        parkinsons & kmeans, k=8 & -0.561$\pm$0.007 \\ 
        parkinsons & HDBSCAN & -0.560$\pm$0.007 \\ 
        temperature & kmeans, k=4 & -0.721$\pm$0.006 \\ 
        temperature & kmeans, k=8 & -0.721$\pm$0.006 \\ 
        temperature & HDBSCAN & -0.722$\pm$0.006 \\ 
        air & kmeans, k=4 & -0.621$\pm$0.006 \\ 
        air & kmeans, k=8 & -0.621$\pm$0.006 \\ 
        air & HDBSCAN & -0.621$\pm$0.006 \\ 
        skillcraft & kmeans, k=4 & -1.577$\pm$0.017 \\ 
        skillcraft & kmeans, k=8 & -1.576$\pm$0.017 \\ 
        skillcraft & HDBSCAN & -1.577$\pm$0.017 \\ 
        \hline
\end{tabular}
    \label{tab:nxbin}
\end{center}

\end{table}

The average test log likelihood on these datasets with different values of $c_0, \gamma$ is shown in Table~\ref{tab:c0gamma}. For this comparison, $\eta=0.1$, maximum depth of trees is 6 for Logistic Regression and reduced to 4 for MLP. The datasets are not rotated. 

\begin{table}[h]
    \centering
\begin{tabular}{lllllllllllllllll}
    \hline
        $c_0$ & $\gamma$ & Energy & Parkinsons & Temperature & Air & Skillcraft  \\ \hline
        0.05 & 0.5 & 1.48$\pm$0.06 & -0.76$\pm$0.01 & -0.80$\pm$0.01 & -0.78$\pm$0.01 & -1.88$\pm$0.02 \\ 
        0.05 & 0.1 & 1.36$\pm$0.06 & -0.76$\pm$0.01 & -0.84$\pm$0.01 & -0.92$\pm$0.01 & -2.16$\pm$0.02\\ 
        0.1 & 0.1 & 1.18$\pm$0.06 & -0.83$\pm$0.01 & -0.86$\pm$0.01 & -0.96$\pm$0.01 & -2.30$\pm$0.02 \\ 
        \hline
    \end{tabular}
    \caption{Average test log-likelihood (mean$\pm$SE) of UCI datasets under different $c_0,\gamma$}
    \label{tab:c0gamma}
\end{table}

\subsection{Additional figures}
\label{a:fig}
This section contains additional figures for the experiments. Specifically, 
the ground truth and estimated density for the simulation examples are provided in ~\ref{fig:toy1}-\ref{fig:toy4}. 

\begin{figure}[h]
  \centering
  \subfigure[Ground truth]{
    \includegraphics[width=0.29\columnwidth]{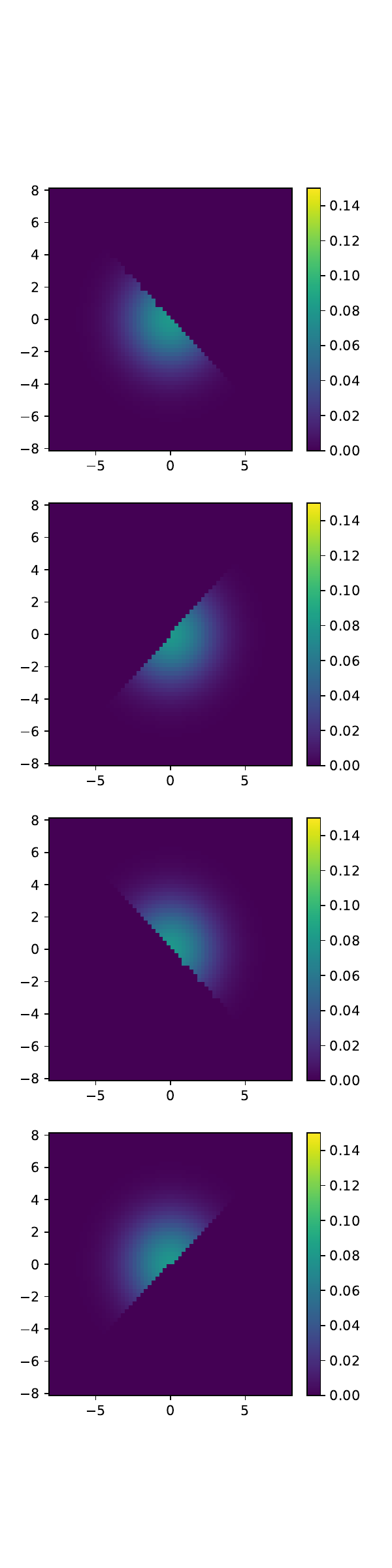}
    \label{fig:subfig1}
  }
    \hfill
  \subfigure[Estimated density (no rotations)]{
    \includegraphics[width=0.29\columnwidth]{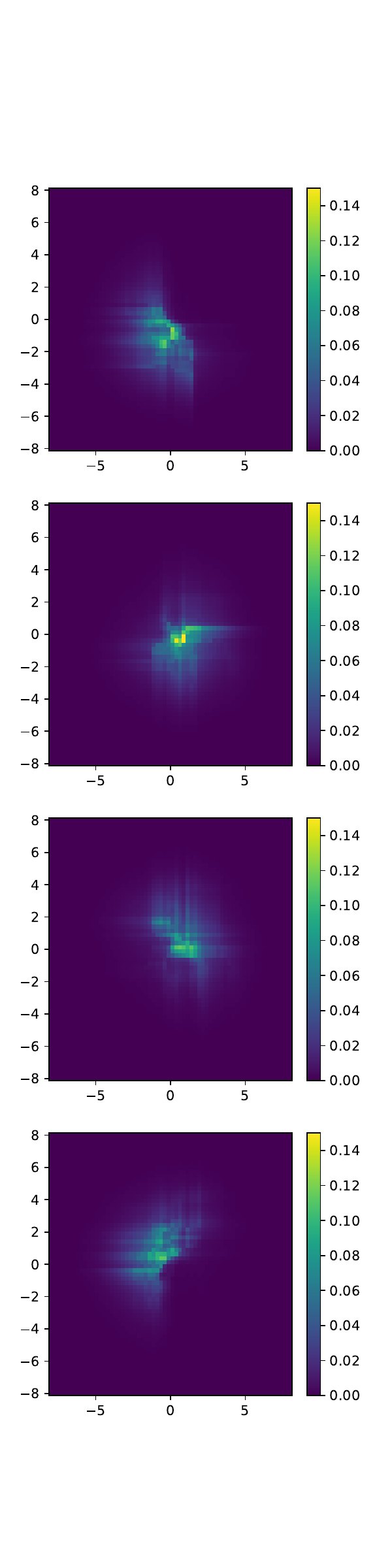}
    \label{fig:subfig3}
  }
  \hfill
  \subfigure[Estimated density (with rotations)]{
    \includegraphics[width=0.29\columnwidth]{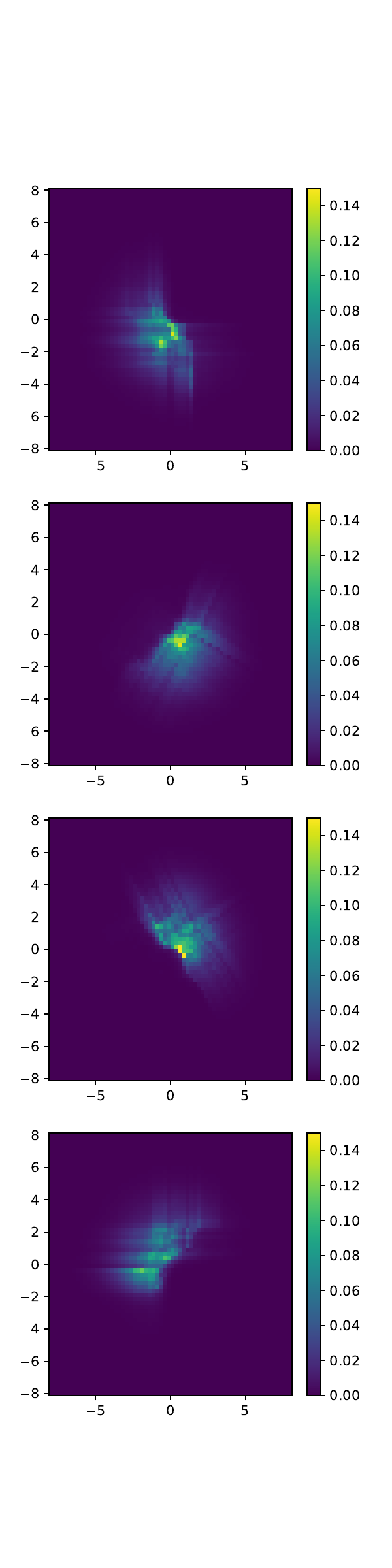}
    \label{fig:subfig2}
  }
  \caption{Half Gaussian. From top to bottom, the rows correspond to $x=-0.75,-0.25, 0.25, 0.75$ respectively. }
  \label{fig:halfgaussian}
\end{figure}

\begin{figure}[h]
  \centering
  \subfigure[Ground truth]{
    \includegraphics[width=0.29\columnwidth]{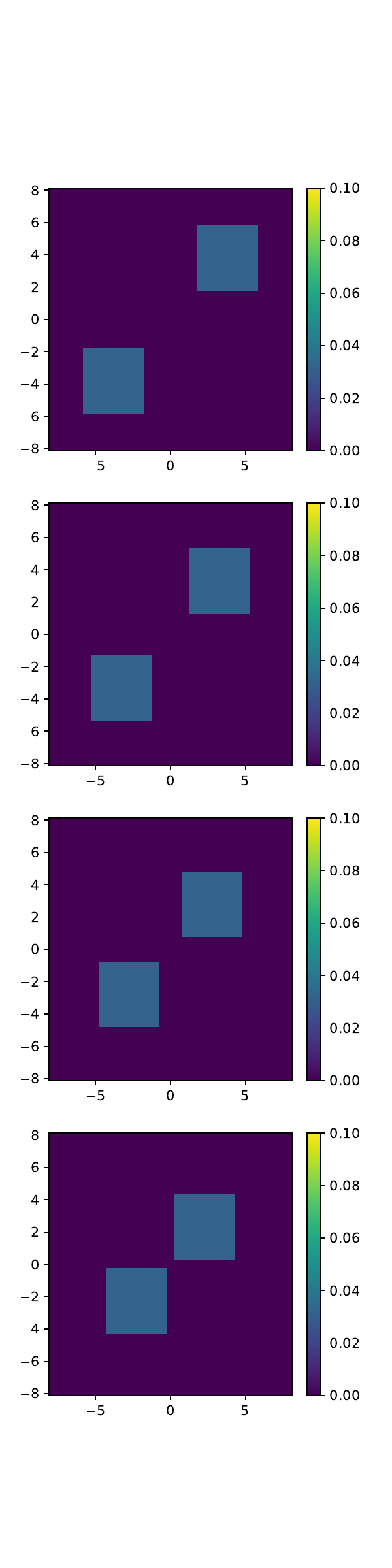}
    \label{fig:subfig1}
  }
    \hfill
  \subfigure[Estimated density (no rotations)]{
    \includegraphics[width=0.29\columnwidth]{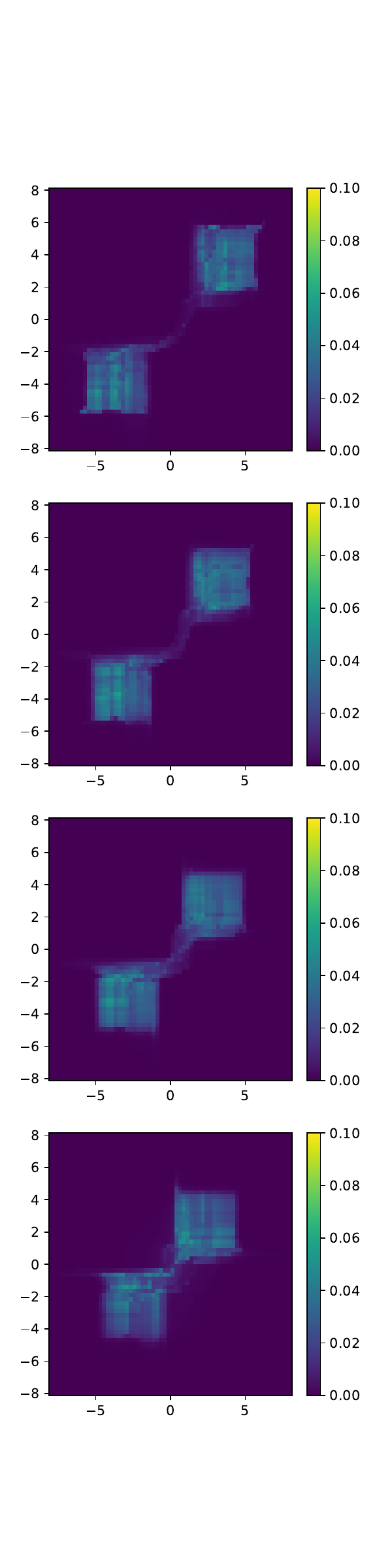}
    \label{fig:subfig3}
  }
  \hfill
  \subfigure[Estimated density (with rotations)]{
    \includegraphics[width=0.29\columnwidth]{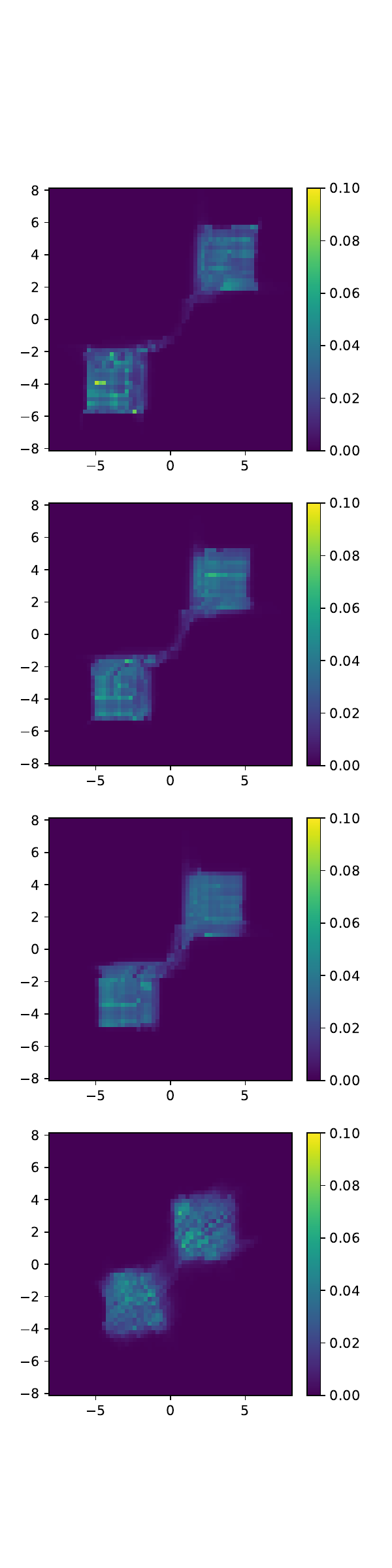}
    \label{fig:subfig2}
  }
  \caption{Squares. From top to bottom, the rows correspond to $x=-0.75,-0.25, 0.25, 0.75$ respectively. }
  \label{fig:toy1}
\end{figure}

\begin{figure}[h]
  \centering
  \subfigure[Ground truth]{
    \includegraphics[width=0.29\columnwidth]{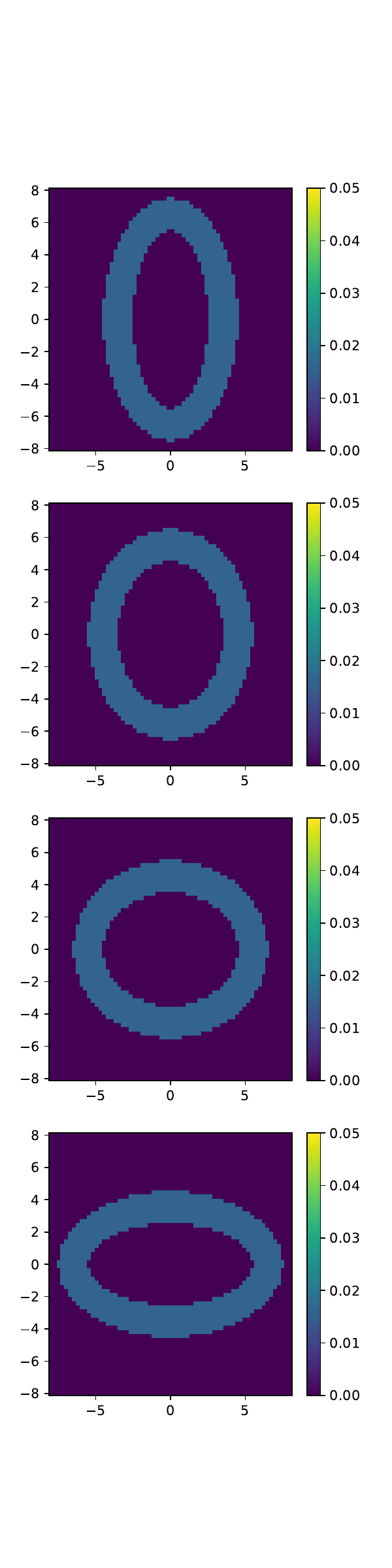}
    \label{fig:subfig1}
  }
    \hfill
  \subfigure[Estimated density (no rotations)]{
    \includegraphics[width=0.29\columnwidth]{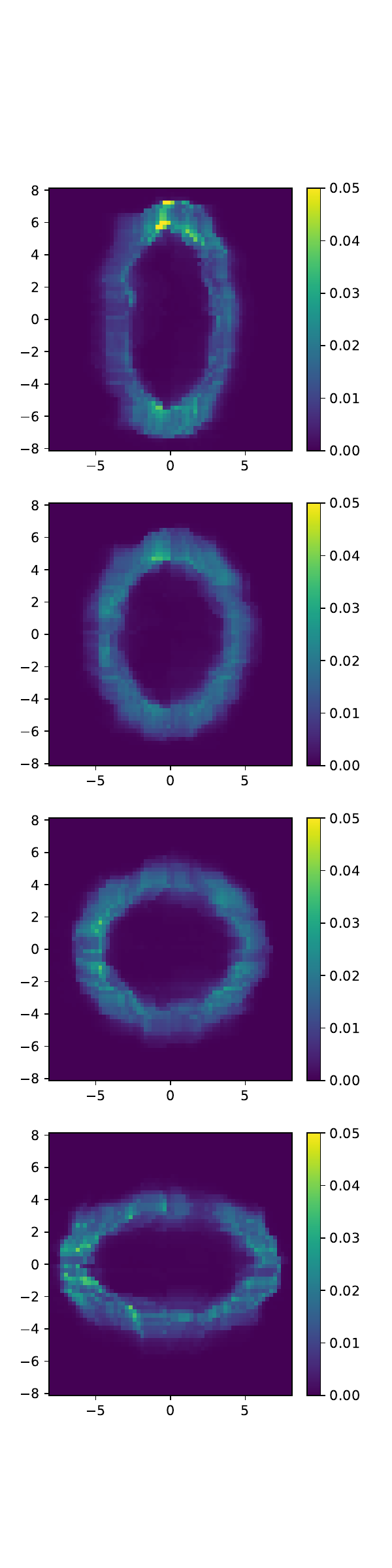}
    \label{fig:subfig3}
  }
  \hfill
  \subfigure[Estimated density (with rotations)]{
    \includegraphics[width=0.29\columnwidth]{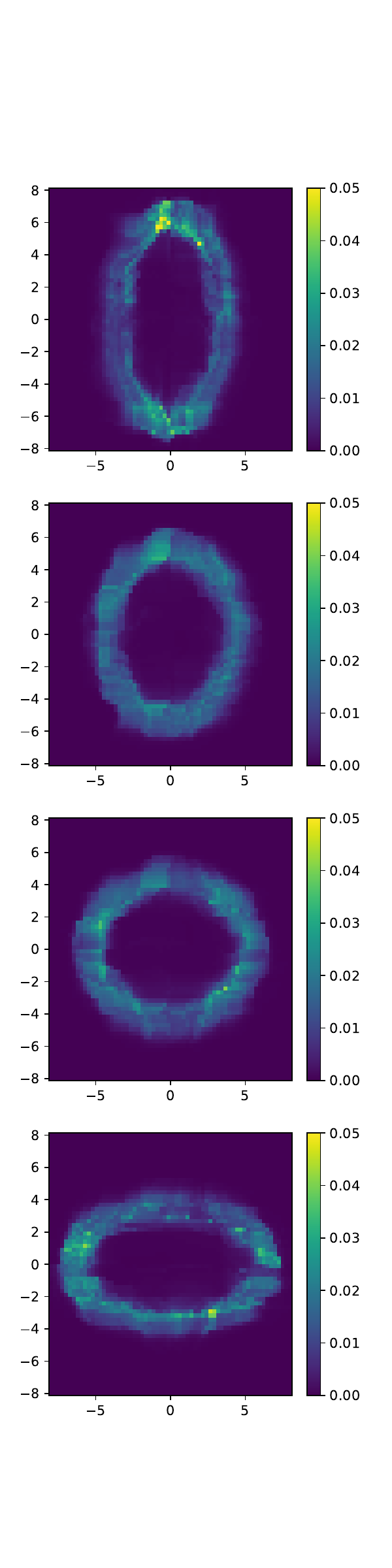}
    \label{fig:subfig2}
  }
  \caption{Elastic Ring. From top to bottom, the rows correspond to $x=-0.75,-0.25, 0.25, 0.75$ respectively. }
  \label{fig:toy3}
\end{figure}

\begin{figure}[h]
  \centering
  \subfigure[Ground truth]{
    \includegraphics[width=0.29\columnwidth]{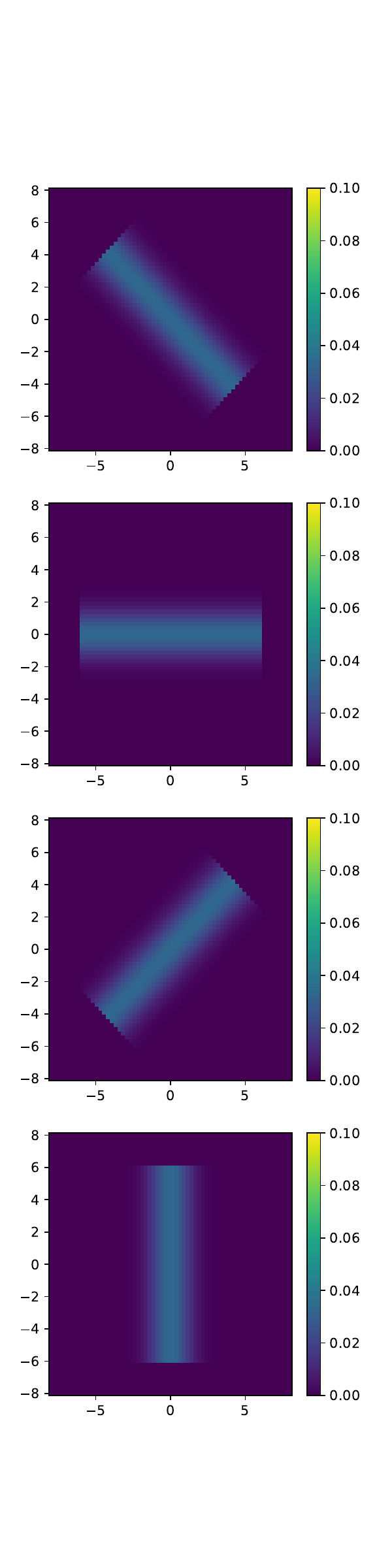}
    \label{fig:subfig1}
  }
    \hfill
  \subfigure[Estimated density (no rotations)]{
    \includegraphics[width=0.29\columnwidth]{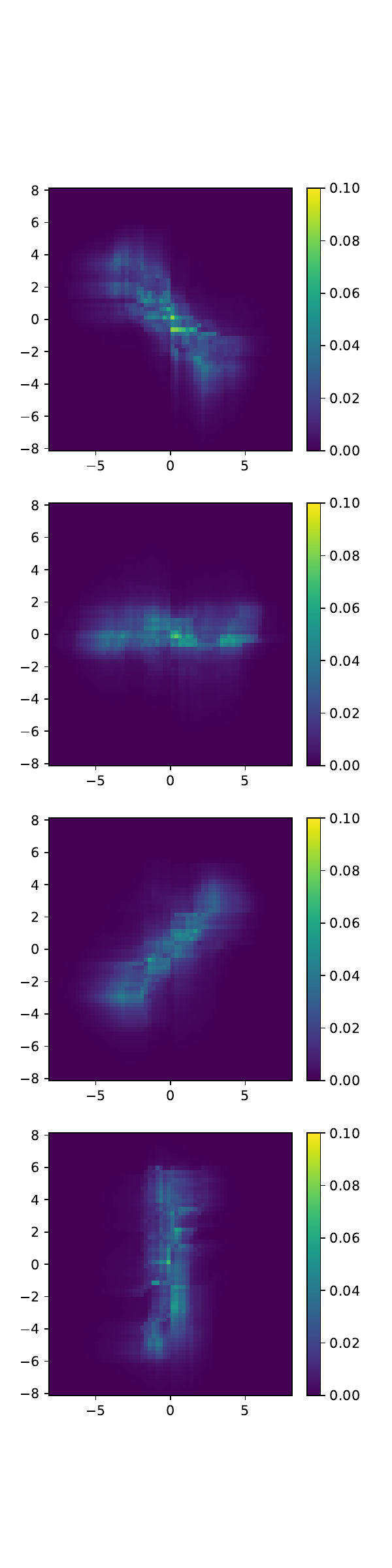}
    \label{fig:a}
  }
  \hfill
  \subfigure[Estimated density (with rotations)]{
    \includegraphics[width=0.29\columnwidth]{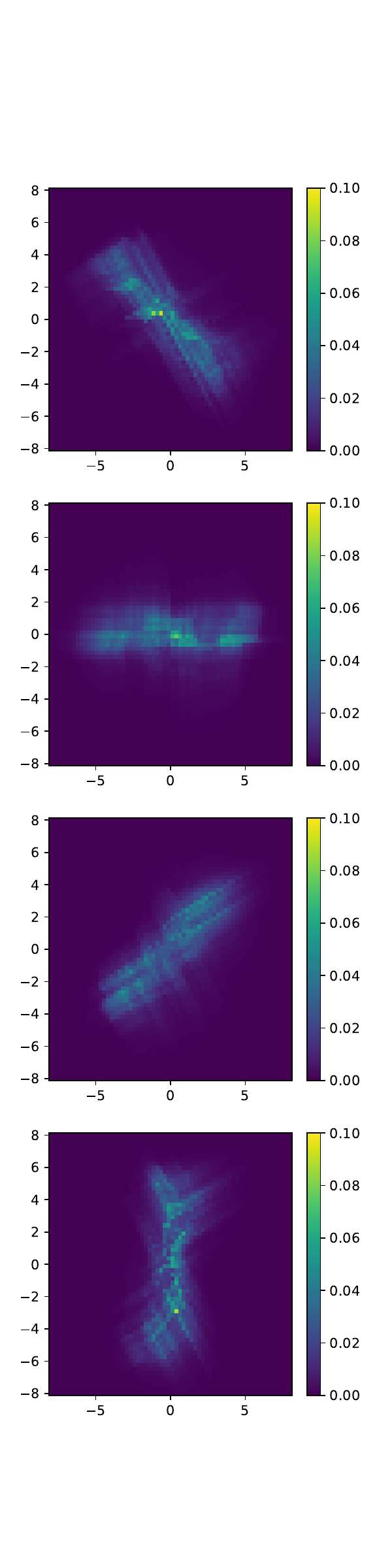}
    \label{fig:subfig2}
  }
  \caption{Gaussian stick. From top to bottom, the rows correspond to $x=-0.75,-0.25, 0.25, 0.75$ respectively. }
  \label{fig:toy4}
\end{figure}

\end{document}